\documentclass{article}

\usepackage{moresize}

\usepackage{natbib}

\usepackage[utf8]{inputenc} 
\usepackage[T1]{fontenc}    
\usepackage{hyperref}       
\usepackage{url}            
\usepackage{booktabs}       
\usepackage{amsfonts}       
\usepackage{nicefrac}       
\usepackage{microtype}      
\usepackage{xcolor}         
\usepackage{algorithm}

\usepackage{amsmath}
\usepackage{amssymb}
\usepackage{mathtools}
\usepackage{amsthm}

\usepackage{algorithm}
\usepackage{algorithmic}
\usepackage[capitalize,noabbrev]{cleveref}

\theoremstyle{plain}
\newtheorem{theorem}{Theorem}[section]

\theoremstyle{definition}

\theoremstyle{remark}

\usepackage[textsize=tiny]{todonotes}

\usepackage{authblk}

\usepackage{microtype}
\usepackage{graphicx}
\usepackage{subfigure}
\usepackage{booktabs} 

\usepackage{amsmath}
\usepackage{amsthm}
\usepackage{dsfont}
\usepackage{amssymb}
\usepackage{amscd}
\usepackage{mathtools}

\usepackage{enumerate}
\usepackage{ bbold }

\usepackage{xcolor}

\newcommand{\argmax}{\operatorname*{argmax}} 
\newcommand{\argmin}{\operatorname*{argmin}}

\newcommand{\actions}{\mathcal{A}}
\newcommand{\un}{\mathbb{1}}
\newcommand{\E}{\mathbb{E}}

\usepackage{numprint}

\title{Bootstrapping Expectiles in Reinforcement Learning}

\author{%
  Pierre Clavier$^{1,2}$, Emmanuel Rachelson$^3$, Erwan Le Pennec$^1$, Matthieu Geist$^4$
     \\
 $^1$CMAP, Ecole Polytechnique, Paris,\\
  $^2$INRIA Paris, HeKA,\\
  $^3$SUPAERO, Toulouse\\
  $^4$Cohere.  \\}
  
\begin{document}

\maketitle

\begin{abstract}
  Many classic Reinforcement Learning (RL) algorithms rely on a Bellman operator, which involves an expectation over the next states, leading to the concept of bootstrapping. To introduce a form of pessimism, we propose to replace this expectation with an expectile. In practice, this can be very simply done by replacing the $L_2$ loss with a more general expectile loss for the critic. Introducing pessimism in RL is desirable for various reasons, such as tackling the overestimation problem (for which classic solutions are double Q-learning or the twin-critic approach of TD3) or robust RL (where transitions are adversarial). We study empirically these two cases. For the overestimation problem, we show that the proposed approach, \texttt{ExpectRL}, provides better results than a classic twin-critic. On robust RL benchmarks, involving changes of the environment, we show that our approach is more robust than classic RL algorithms. We also introduce a variation of \texttt{ExpectRL} combined with domain randomization which is competitive with state-of-the-art robust RL agents. Eventually, we also extend \texttt{ExpectRL} with a mechanism for choosing automatically the expectile value, that is the degree of pessimism. \looseness=-1
\end{abstract}
\section{Introduction}
Pessimism is a desirable concept in many Reinforcement Learning (RL) algorithms to stabilize the learning and get an accurate estimation of the value function. This idea is developed in Double Q-learning  \citep{hasselt2010double}, an RL technique designed to address the issue of overestimation bias in value estimation, a common challenge in Q-learning and related algorithms. Overestimation bias occurs when the estimated values of actions are higher than their true values, potentially leading to a suboptimal policy.  By maintaining two sets of Q-values and decoupling action selection from value estimation, Double Q-learning provides a more accurate and less optimistic estimate of the true values of actions. In general, Double Q-learning enhances the stability of the learning process and these principles can be extended to deep RL known as Double Deep Q-Networks (DDQN), a successful approach in various applications \citep{van2016deep}.
Pessimism also appears in the twin critic approach, the equivalent of Double Q-learning for continuous action spaces,  which requires training two critics to select the most pessimistic one. Many state-of-the-art RL algorithms are based on this method, such as TD3  \citep{fujimoto2018addressing} that uses this method to improve on DDPG \citep{lillicrap2015continuous} and  SAC \citep{haarnoja2018soft} that uses this trick to stabilize the learning of $Q$-functions and policies. \looseness=-1

\vspace{0.3cm}
The idea of pessimism is also central in Robust RL \citep{moos2022robust}, where the agent tries to find the best policy under the worst transition kernel in a certain uncertainty space.  It has been introduced first theoretically in the context of Robust MDPs \citep{iyengar2005robust,nilim2005robust} (RMDPs) where the transition probability varies in an uncertainty (or ambiguity) set. Hence, the solution of robust MDPs is less sensitive to model estimation errors with a properly chosen uncertainty set, as RMDPs are formulated as a max-min problem, where the objective is to find the policy that maximizes the value function for the worst possible model that lies within an uncertainty set around a nominal model.  Fortunately, many structural properties of MDPs are preserved in RMDPs \citep{iyengar2005robust}, and methods such as robust value iteration, robust modified policy iteration, or partial robust policy iteration \citep{ho2021partial}  can be used to solve them. It is also known that the uncertainty in the reward can be easily tackled while handling uncertainty in the transition kernel is much more difficult \citep{kumar2022efficient,derman2021twice}. Finally, the sample complexity of RMDPs has been studied theoretically \citep{yang2021towards,shi2022distributionally, clavier2023towards,shi2023curious}. However, these works usually assume having access to a generative model.  \looseness=-1

\vspace{0.3cm}

Robust RL \citep{moos2022robust} tries to bridge a gap with real-life problems, classifying its algorithms into two distinct groups. The first group engages solely with the nominal kernel or the center of the uncertainty set. To enhance robustness, these algorithms often adopt an equivalent risk-averse formulation to instill pessimism. For instance, \citet{clavier2022robust} employ mean-standard deviation optimization through Distributional Learning to bolster robustness. Another strategy involves introducing perturbations on actions during the learning process, as demonstrated by \citet{tessler2019action}, aiming to fortify robustness during testing. Another method, known as adversarial kernel robust RL \citep{wang2023robust}, exclusively samples from the nominal kernel and employs resampling techniques to simulate the adversarial kernel. While this approach introduces a novel paradigm, it also leads to challenges associated with poor sample complexity due to resampling and requiring access to a generative model. Despite this drawback, the adversarial kernel robust RL paradigm offers an intriguing avenue for exploration and development in the realm of robust RL. Finally, policy gradient \citep{kumar2023policy,li2023policy} in the case of Robust MDPs is also an alternative. A practical algorithm using robust 
 policy gradient with Wasserstein metric is proposed by \citet{abdullah2019wasserstein}, but this approach requires having access to model parameters which are usually not available in a model-free setting. The second category of algorithms engages with samples within the uncertainty set, leveraging available information to enhance the robustness and generalization of policies to diverse environments. Algorithms within this category, such as IWOCS \citep{zouitine2023revisiting}, M2TD3 \citep{tanabe2022max}, M3DDPG \citep{li2019robust}, and RARL \citep{pinto2017robust} actively interact with various close environments to fortify robustness in the context of RL.  \looseness=-1

 \vspace{0.3cm}

In all these settings, the idea of pessimism is central. We propose here a new simple form of pessimism based on expectile estimates that can be plugged into any RL algorithm. For a given algorithm, the only modification relies on the critic loss in an actor-critic framework or in the $Q$-learning loss for $Q$-function based algorithms. Given a target $ y\left(r, s^{\prime} \right)=r+\gamma Q_{\phi_{, \operatorname{targ}}}\left(s^{\prime}, \pi(s^{\prime})\right)$ with reward $r$, policy $\pi$, we propose to minimize
\begin{align*}
    L\left(\phi_, \mathcal{D}\right)=\underset{\left(s, a, r, s^{\prime}\right) \sim \mathcal{D}}{\mathbb{E}}\left[L^\alpha_2\left(Q_{\phi}(s, a)-y\left(r, s^{\prime}\right)\right)\right],
\end{align*}
where $L_2^\tau$ is the expectile loss defined in Section \ref{exextile}. For $\alpha=1/2$, the expectile coincides with the classical mean, and we recover the classical $L_2$ loss of most RL algorithms. We denote this modification as \texttt{ExpectRL}. In many RL algorithms, we are bootstrapping the expectation of the $Q$-function over the next state, by definition of the classical Bellman equation.

\vspace{0.3cm}

Our method \texttt{ExpectRL} is equivalent to bootstrapping the expectile and not the expectation of the Q value. Bootstraping expectiles still leads to an algorithm with the contraction mapping property for the associated Expectile Bellman Operator, but adds pessimism by giving more weight to the pessimistic next state compared to a classical expectation (see Section~\ref{exextile}).
\looseness=-1

The \texttt{ExpectRL} modification is relevant in the context of the twin critic approach as when employing this method, the challenge arises in effectively regulating the level of pessimism through the application of the twin critic method, which remains heuristic for continuous action spaces, although it has been studied in the discrete case by \citet{hasselt2010double}. Furthermore, the acquisition of imprecise $Q$ functions has the potential to yield detrimental outcomes in practical applications, introducing the risk of catastrophic consequences. Using the \texttt{ExpectRL} method,  the degree of pessimism in learning the value or $Q$ function is controlled through the parameter $\alpha$,
and our first question is: \looseness=-1
\vspace{0.3cm}

\textit{Can we replace the learning of two critics in the twin critic method, using only a simple expectile bootstrapping? 
}

\vspace{0.3cm}

In the Robust RL setting, \texttt{ExpectRL} can also be beneficial as by nature expectiles are a coherent, convex risk measure, that can be written as a minimum of an expectation over probability measure on a close convex set \citep{delbaen2002coherent}. So implicitly bootstrapping an expectile instead of an average leads to a robust RL algorithm. Compared to many Robust RL algorithms, our method is simple in the sense that the $\alpha$-expectile is more interpretable and easy to choose than a penalization or trade-off parameter in mean-standard deviation optimization \citep{clavier2022robust}. 
\texttt{ExpectRL} has the advantage of being computationally simple compared to other methods,  as it uses all samples, compared to the work of \citet{wang2023robust}, that needs resampling to induce robustness. Finally, our method is simple and can be adapted to practical algorithms, compared to robust policy gradient methods such as \cite{kumar2023policy,li2023policy}.
Moreover, while these algorithms can be considered more mathematically grounded and less heuristic, the second group with IWOCS, M2DTD2, RARL 
\citep{zouitine2023revisiting,tanabe2022max,li2019robust,pinto2017robust} tends to rely on heuristic approaches that exhibit practical efficacy on real-world benchmarks.
This dichotomy prompts the question:

\vspace{0.3cm}

\textit{Can we leverage \texttt{ExpectRL} method as a surrogate for Robust RL and formulate robust RL algorithms that are both mathematically founded and requiring minimal parameter tuning?}

\vspace{0.3cm}

By extending expectile bootstrapping (\texttt{ExpectRL}) with sampling from the entire uncertainty set using domain randomization (DR), our approach 
bolsters robustness, positioning itself competitively against the best-performing algorithms. Notably, our algorithm incurs low computational costs relatively to other algorithms and requires minimal or no hyperparameter tuning. 
Our contributions are the following. \looseness=-1

\vspace{0.3cm}

 Our \textbf{first contribution}, is to introduce \texttt{ExpectRL}, and demonstrate the efficacy of that method as a viable alternative to the twin critic trick with $L_2$ loss across diverse environments. This substitution helps empirically control of the overestimation in the $Q$-function, thereby reducing the computational burden associated with the conventional application of the twin trick, which entails learning two critics.

 \vspace{0.3cm}

    The \textbf{ second contribution} of our work lies in establishing that expectile bootstrapping or \texttt{ExpectRL} facilitates the development of straightforward Deep Robust RL approaches. These approaches exhibit enhanced robustness compared to classical RL algorithms. The effectiveness of our approach combining \texttt{ExpectRL} with DR is demonstrated 
 on various benchmarks and results in an algorithm that closely approaches the state of the art in robust RL, offering advantages such as lower computational costs and minimal hyperparameters to fine-tune.

 \vspace{0.3cm}
 Our \textbf{third contribution} introduces an algorithm, \texttt{AutoExpectRL} that leverages an automatic mechanism for selecting the expectile or determining the degree of pessimism. Leveraging bandit algorithms, this approach provides an automated and adaptive way to fine-tune the expectile parameter, contributing to the overall efficiency and effectiveness of the algorithm. \looseness=-1

\section{Related Work}
\paragraph{TD3 and twin critics.}

To tackle the problem of over-estimation of the value function, TD3 algorithm  \citep{fujimoto2018addressing} algorithm uses two critics. Defining the target $y_{min}$ as \looseness=-1
$y_{min}\left(r, s^{\prime} \right)=r+\gamma \min _{i=1,2} Q_{\phi_{i, \operatorname{targ}}}\left(s^{\prime}, \pi\left(s^{\prime}\right)\right),$
both critics are learned by regressing to this target, such that, for $i \in\{1,2\}$,
$L\left(\phi_i, \mathcal{D}\right)=\underset{\left(s, a, r, s^{\prime}, d\right) \sim \mathcal{D}}{\mathbb{E}}\left(Q_{\phi_i}(s, a)-y_{min}\left(r, s^{\prime}\right)\right)^2 $
Our approach is different as we do not consider the classic $L_2$ loss and only use one critic. We will compare \texttt{{ExpectRL}} to the classic TD3 algorithm both with twin critics and one critic to understand the influence of our method. \looseness=-1
\paragraph{Expectiles in Distributional RL.}
Expectiles have found application within the domain of Distributional RL (RL), as evidenced by studies such as \citep{rowland2019statistics,dabney2018implicit,jullien2023distributional}. It is crucial to note a distinction in our approach, where we specifically focus on learning a single expectile to substitute the conventional $L_2$ norm. This diverges from the methodology adopted in these referenced papers, where the entire distribution is learned using different expectiles. Moreover, they do not consider expectile statistics on the same random variable as they consider expectiles of the full return.\looseness=-1
\paragraph{Expectile in Offline RL and the IQL algorithm .}
Implicit Q-learning (IQL) \citep{kostrikov2021offline} in the context of offline RL endeavors to enhance policies without the necessity of evaluating actions that have not been encountered. Like our method, IQL employs a distinctive approach by treating the state value function as a random variable associated with the action, but achieves an estimation of the optimal action values for a state by utilizing a state conditional upper expectile.
In \texttt{{ExpectRL}}, we employ lower expectiles to instill pessimism on the next state and approximate a minimum function, contrasting with the conventional use of upper expectiles for approximating the maximum in the Bellman optimality equation.\looseness=-1
\paragraph{ Risk-Averse RL.}
Risk-averse RL, as explored in studies like \cite{pan2019risk}, diverges from the traditional risk-neutral RL paradigm. Its objective is to optimize a risk measure associated with the return random variable, rather than focusing solely on its expectation. Within this framework, Mean-Variance Policy Iteration has been considered for optimization, as evidenced by \citet{zhang2021mean}, and Conditional Value at Risk ($\mathrm{CVaR}$), as studied by \citet{greenberg2022efficient}.
The link between Robust and Risk averse MDPS has been highlighted by 
\citet{chow2015risk} and \citet{zhang2023regularized} who provide a mathematical foundation for risk-averse RL methodologies, emphasizing the significance of coherent risk measures in achieving robust and reliable policies. Our method lies in risk-averse RL as expectiles are a coherent risk measure \citep{zhang2023regularized}, but to the best of our knowledge, the expectile statistic has never been considered before for tackling robust RL problems. \looseness=-1
\paragraph{Regularisation and robustness in RL.}
Regularization plays a pivotal role in the context of Markov Decision Processes (MDPs), as underscored by \citet{derman2021twice} or \citet{eysenbach2021maximum}, who have elucidated the pronounced connection between robust MDPs and their regularized counterparts. Specifically, they have illustrated that a regularised policy during interaction with a given MDP exhibits robustness within an uncertainty set surrounding the MDP in question. In this work, we focus on the idea that generalization, regularization, and robustness are strongly linked in RL or MDPs as shown by \citet{husain2021regularized,derman2020distributional,derman2021twice,ying2021towards,brekelmans2022your}. The main drawback of this method is that it requires tuning the introduced penalization to improve robustness, which is not easy in practice as it is very task-dependent. The magnitude of the penalization is not always interpretable compared to $\alpha$, the value of the expectile.\looseness=-1
\section{Background}
\subsection{ Markov Decision Processes }
We first define Robust Markov Decision Processes (MDPs) as $\mathcal{M}_{\Omega}=\left\{\mathcal{M}_\omega\right\}_{\omega \in \Omega}$, with $\mathcal{M}_\omega=\left\langle S, A, P_\omega, P_\omega^0, r_\omega, \gamma\right\rangle$ the MDP with specific uncertainty parameter $\omega \in \Omega$. The chosen state space $S$ and action space $A$  are subsets of real-valued vector spaces in our setting. The transition probability density $P_\omega: S \times A \times S \rightarrow \mathbb{R}$, the initial state probability density $P_\omega^0: S \rightarrow \mathbb{R}$, and the immediate reward $r_\omega: S \times A \rightarrow \mathbb{R}$ depend on $\omega$. 
Moreover, we define $P_{sa,w}$ the vector of  $P_\omega(s,a,.)$.
The discount factor is denoted by $\gamma \in(0,1)$.
Let $\pi_\theta: S \rightarrow A$ be a  policy parameterized by $\theta \in \Theta$ and $\pi^*$ the optimal policy. Given an uncertainty parameter $\omega \in \Omega$, the initial state follows $s_0 \sim P_\omega^0$. At each time step $t \geqslant 0$, the agent observes state $s_t$, selects action $a_t=\pi_\theta\left(s_t\right)$, interacts with the environment, and observes the next state $s_{t+1} \sim P_\omega\left(\cdot \mid s_t, a_t\right)$,  and the immediate reward $r_t=r_\omega\left(s_t, a_t\right)$. 
The discounted return of the trajectory starting from time step $t$ is $R_t=\sum_{k \geqslant 0} \gamma^k r_{t+k}$.
The action value function $q^{\pi_\theta}(s, a, \omega)$  and optimal action value  $q^{*}(s, a, \omega)$ under $\omega$ is the expectation of $R_t$ starting with $s_t=s$ and $a_t=$ $a$ under $\omega$; that is, \looseness=-1 
\begin{align*}
   &q^{\pi_\theta}(s, a, \omega)=\mathbb{E}_{P_\omega,\pi_\theta}\left[R_t \mid s_t=s, a_t=a \right] , \quad
   q^{*}(s, a, \omega)=\mathbb{E}_{P_\omega,\pi^*}\left[R_t \mid s_t=s, a_t=a \right] ,
\end{align*}
where $\mathbb{E}$ is the expectation. Note that we introduce $\omega$ to the argument to explain the $Q$-value dependence on $\omega$. Lastly, we define the value function as 
\begin{align*}
   &v^{\pi_\theta}(s, \omega)=\mathbb{E}_{P_\omega,\pi_\theta}\left[R_t \mid s_t=s \right], \quad v^{*}(s, \omega)=\mathbb{E}_{P_\omega,\pi^*}\left[R_t \mid s_t=s \right].
\end{align*}
In the following, we will drop the $\omega$ subscript for simplicity and define the expectile (optimal) value function, that follows the recursive Bellman equation \looseness=-1
\begin{align}
  v^{\pi,P}(s) &=  v^\pi(s) = \E_{a\sim\pi(\cdot|s)}[\underbrace{r(s,a) + \gamma \mathbb{E}_{P_{sa}}  [ v^\pi]}_{\triangleq q^\pi(s,a)}],
    \\
    v^*(s) &= \max_{a\in\actions}(\underbrace{r(s,a) + \gamma \mathbb{E}_{P_{sa}}[ v^*]}_{\triangleq q^*(s,a)})).
\end{align}
Finally, we define the classical Bellman Operator  and optimal Bellman Operator that are $\gamma$-contractions, so iteration of these operators leads $v^\pi$ or $v^*$:
\begin{align}
   & (T^\pi v)(s) := (T^\pi_{r,P} v)(s)= \sum_a \pi(a|s) (r(s,a) + \gamma \mathbb{E}_{P_{sa}}[v]) \\
   & (T^* v)(s) :=  (T^{\pi^*}_{r,P} v)(s) = \max _a (r(s,a) + \gamma \mathbb{E}_{P_{sa}}[v]).
\end{align}

\subsection{Robust MDPs}
Once classical MDPs are defined, we can define robust (optimal) Bellman operators $\mathcal{T}^\pi_{\mathcal{U}}$ and  $\mathcal{T}_{\mathcal{U}}^*$,
\begin{align}
    (T^\pi_{\mathcal{U}}v)(s):=\min _{r, P \in \mathcal{U}}\left(T_{r, P}^\pi v\right)(s),
    \left(T_{\mathcal{U}}^* v\right)(s):=\max _{\pi \in \Delta_{\mathcal{A}}} \min _{r, P \in \mathcal{U}}\left(T_{r, P}^{\pi} v\right)(s),
\end{align}
where $P$ and $r$ belong  to the uncertainty set $\mathcal{U}$. The optimal robust Bellman operator $T_{\mathcal{U}}^*$ and robust Bellman operator $T^\pi_{\mathcal{U}}$ are $\gamma$-contraction maps for any policy $\pi$ \citep[Thm.~3.2]{iyengar2005robust} if  the uncertainty set $\mathcal{U}$ is a subset of $\Delta_s$ where $\Delta_s$ it the simplex of $\vert S\vert $ elements so that the transition kernel is valid.
%
Finally, for any initial values $v_0^\pi, v_0^*$, sequences defined as
$
v_{n+1}^\pi:=T_{\mathcal{U}}^\pi v_n^\pi$ and $v_{n+1}^*:=T_{\mathcal{U}}^* v_n^*
$
converge linearly to their respective fixed points, that is $v_n^\pi \rightarrow v_{\mathcal{U}}^\pi$ and $v_n^* \rightarrow v_{\mathcal{U}}^*$. \looseness=-1
\subsection{Expectiles}
\label{exextile}
Let's first define expectiles. For $\alpha\in(0,1)$ and $X$  a random variable, the $\alpha$-expectile is defined as $m_\alpha(X) = \argmin_{m} \E_{x}[L_2^\alpha (x - m)] $ with $  L_2^\alpha(u) = |\alpha - \un_{\{u<0\}}|u^2 = \alpha u_+^2 + (1-\alpha) u_-^2,$
where $u_+ = \max(u,0)$ and $u_- = \max(-u,0)$. We can recover the classical mean with $m_{\frac 1 2}(X) = \E[X]$ as $L_2^{1/2}(u) = u^2$.\looseness=-1 Expectiles are gaining interest in statistics and finance as
they induce the only law-invariant, coherent \citep{artzner1999coherent} and elicitable \citep{gneiting2011making} risk measure.  Using the coherent property representation \citep{delbaen2000draft}, one has that $\rho: L^{\infty} \rightarrow \mathbb{R}$ is a coherent risk measure  if and only if there exists a closed convex set $\mathcal{P}$ of $P$-absolutely continuous probability measures such that  $\rho(X)=\inf _{Q \in \mathcal{P}} \mathbb{E}_Q[X], \forall X \in L^{\infty}.$ with $L^{\infty}$ the vector space of essentially bounded measurable functions with the essential supremum norm.  The uncertainty set induced by expectiles as been described by \cite{delbaen2013remark} as $m_\alpha(X)=\min _{Q \in \mathcal{E}}\mathbb{E}_{Q}[X]$ such as
\begin{align}
     \mathcal{E}=\left\{ Q \in \mathcal{P} \mid\  \exists \eta>0,  \eta \sqrt{\frac{\alpha}{1-\alpha}}\leq \frac{d Q}{d P} \leq \sqrt{\frac{(1-\alpha)}{\alpha}} \eta\right\} 
\end{align}
where we define $\frac{d Q}{d P}$ as the Radon-Nikodym derivative of $Q$ with respect to $P$.  Here, the uncertainty set corresponds thus to a lower and upper bound on $\frac{d Q}{d P}$ with a quantity depending on the degree of uncertainty. For $\alpha=1/2$, the uncertainty set becomes the null set and we retrieve the classical mean. This variational form of the expectile will be useful to link risk-sensitive and robust MDPs formulation in the next section. \looseness=-1
\section{\texttt{ExpectRL} method}
First, we introduce the Expectile Bellman Operator and then we will explain our proposed method \texttt{ExpectRL} and \texttt{AutoExpectRL} that work both in classic and robust cases. \looseness=-1
\subsection{Expectile Bellman Operator}
In this section, we derive the loss and explain our approach.
Recall that for $\alpha\in(0,1)$ and $X$  a random variable taking value $x$ and following a probability law $P$, the $\alpha$-expectile is denoted $m_\alpha(X)$ or $ m_\alpha(P,x)$ in the following. Writing the classical Bellman operator for $q$ function
\begin{align}
    (Tq)(s,a) &= r(s,a) + \gamma \langle P_{sa}, v\rangle \nonumber = r(s,a) + \gamma \E_{s'\sim P_{sa}(\cdot)}[v(s')].
\end{align}
and denoting  $V_{sa}$ the random variable which is equal to $v(s')$ with probability $P_{sa}(s')$, it holds that:
\begin{align}
    (Tq)(s,a) &= r(s,a) + \gamma m_{\frac {1}{2}}(V_{sa}) \nonumber = r(s,a) + \gamma m_{\frac {1}{2}}(P_{sa},v) . 
\end{align}
Our method consists instead in considering the following Expectile Bellman operator
\begin{equation}
    (T_\alpha q)(s,a) = r(s,a) + \gamma m_{\alpha}(V_{sa}).
    \label{eq:exp_robust}
\end{equation}
With $\alpha < \frac 1 2$, Eq.~\eqref{eq:exp_robust}  allows to learn a robust policy, in the sense that it is a pessimistic estimate about the value we bootstrap according to the value sampled according to the nominal kernel.
 Next, we define the expectile value of a given policy and the optimal expectile value as:
\begin{align}
    v^\pi_\alpha(s) &= \E_{a\sim\pi(\cdot|s)}[\underbrace{r(s,a) + \gamma m_\alpha(P_{sa}, v^\pi_\alpha)}_{\triangleq q^\pi_\alpha(s,a)}],
    \quad \\
    v^*_\alpha(s) &= \max_{a\in\actions}(\underbrace{r(s,a) + \gamma m_\alpha(P_{sa}, v^*_\alpha}_{\triangleq q^*_\alpha(s,a)})).
\end{align}
With $\alpha=\frac{1}{2}$, we retrieve the standard Bellman equations but we consider  $\alpha < \frac{1}{2}$ for the robust case. Finally, we define (optimal) expectile Bellman Operator as:
    \begin{align*}
    &(T^\pi_\alpha v)(s) = \sum_a \pi(a|s) (r(s,a) + \gamma m_\alpha(P_{sa},  v)). \\
    &(T^*_\alpha v)(s) = \max_a  (r(s,a) + \gamma m_\alpha(P_{sa}, v)). 
   \end{align*}
\begin{theorem}{}
The (optimal) Expectile Bellman Operators are $\gamma$-contractions for the sup norm. (proof in Appx. \ref{contraction}).
    \looseness=-1
\end{theorem}
So as $T_\alpha^\pi$  and $T^*_\alpha$ are  $\gamma$-contractions, it justifies the definition of fixed point $v_\tau^\pi$ and  $v^*_\alpha$. 
 The central idea to show that expectile bootstrapping or \texttt{ExpectRL} is implicitly equivalent to Robust RL  comes \citep{zhang2023regularized} where we try to estimate the optimal robust value function  $v^*_{\mathcal{E}}=\max_\pi \min _{Q \in \mathcal{E}} v^{\pi,Q} $.
\begin{theorem}
    The (optimal) Expectile value function is equal to the (optimal) robust value function
    \begin{align}
         &v^*_\alpha(s)=v^\pi_{\mathcal{E}}:=\max_\pi \min _{Q \in \mathcal{E}} v^{\pi,Q} ,  \quad v^\pi_\alpha(s)=v^\pi_{\mathcal{E}}:= \min _{Q \in \mathcal{E}} v^{\pi,Q}   
    \end{align}
\end{theorem}
where $\mathcal{E}$ is defined in \ref{exextile}.  Proof can be found in \ref{appendix:link_robust_sensitive}. Note that his formulation does not converge to the expectile of the value distribution but to $v^*_\mathcal{E}$ the robust value function. Moreoever, for $\alpha>1/2$, Expectile Bellman operator is not anymore a contraction and there is no theoretical convergence guarantees for risk-seeking RL, not considered here. Now that expectile operators are defined, we will define the related loss. \looseness=-1
\subsection{ The \texttt{ExpecRL } Loss }
In this section, we present the method more from a computational and practical point of view. As stated before, this method can be plugged into any RL algorithm where a $Q$-function is estimated, which included any $ Q$-function-based algorithm or some actor-critic framework during the critic learning. For a given algorithm, the only modification relies on modifying the $L_2$ loss in the $Q$-value step by the Expectile loss. Given a target $ y\left(r, s^{\prime} \right)=r+\gamma Q_{\phi_{, \operatorname{targ}}}\left(s^{\prime}, \pi(s^{\prime})\right)$ with reward $r$, policy $\pi$, we propose to minimize\looseness=-1
\begin{align}
    L\left(\phi_, \mathcal{D}\right)=\underset{\left(s, a, r, s^{\prime}\right) \sim \mathcal{D}}{\mathbb{E}}\left[L^\alpha_2\left(Q_{\phi}(s, a)-y\left(r, s^{\prime}\right)\right)\right],
\end{align}
where $L_2^\alpha$ is the expectile loss defined in Section \ref{exextile}. For $\alpha=1/2$, the expectile coincides with the classical mean, and we retrieve the classical $L_2$ loss present in most RL algorithms.
We will use TD3 as a baseline and replace the learning of the critic with this loss. The actor loss remains the same in the learning process. With \texttt{ExpectRL}, only one critic is needed, replacing the double critic present in this algorithm. We will compare our method with the classical TD3 algorithm using the twin critic trick and TD3 with one critic to see the influence of our method.\looseness=-1
\subsection{ \texttt{ExpecRL} method with Domain randomisation }
From a practical point of view, many  Robust RL algorithms such as M2TD3 \citep{tanabe2022max}, M3DDPG \citep{li2019robust}, and RARL \citep{pinto2017robust} not only interact with the nominal environment but also with environments that belong to the uncertainty set $\mathcal{U}$. Sampling trajectories from the entire uncertainty set allows algorithms to get knowledge from dangerous trajectories and allows algorithms to generalize better than algorithms that only sample from the nominal. Receiving information about all environments that need to be robust during the training phase, the algorithm tends to obtain better performance on minimum performance over these environments on testing. With the same idea of generalization, Domain Randomisation (DR) \citep{tobin2017domain} focuses not on the worst case under the uncertainty set but on the expectation. Given a point of the uncertainty set $P_\omega \in \mathcal{U}$,
the DR objective is: \looseness=-1
  $  \pi^*_{\mathrm{DR}} =\argmax _{\pi} \mathbb{E}_{\omega \in \Omega, s \sim P^{0}_\omega}   [  v^\pi(s,w)    ].$
In other words, DR tries to find the best policy on average over all environments in the uncertainty set. The approach we propose to be competitive on a robust benchmark is to find the best policy using \texttt{ExpectRL} under domain randomization or \looseness=-1
\begin{align}
    \pi^*_{\mathrm{DR},\alpha} &=\argmax _{\pi} \mathbb{E}_{\omega \in \Omega, s \sim P^{0}_\omega}   [  v^\pi_\alpha(s,\omega)    ] = \argmax _{\pi} \mathbb{E}_{\omega \in \Omega, s \sim P^{0}_\omega}   [  \min _{P_\omega \in \mathcal{E} } v^{\pi,P}(s,\omega)    ],
\end{align}
where $ v^\pi_\alpha(s,\omega) $ is the  expectile value function  under uncertainty kernel $P_\omega$ and $\mathcal{E}$ defined in Section \ref{exextile}. Using this approach, we hope to get sufficient information from all the environments using DR and improve robustness and worse-case performance using \texttt{ExpectRL}.
The advantage of the approach is that any algorithm can be used for learning the policy, sampling from the entire uncertainty set uniformly and replacing the critic loss of this algorithm learning with \texttt{ExpecRl} loss. The effectiveness of this algorithm on a Robust benchmark will be conducted in Section \ref{benchmark:robust}. Getting an algorithm that is mathematically founded and which tries to get the worst-case performance, the last question is how to choose the degree of pessimism or $\alpha \in (0,1/2)$ in practice. The following section tries to answer this question using a bandit algorithm to auto-tune $\alpha$. \looseness=-1
\subsection{Auto-tuning of the expectile $\alpha$ using bandit  }
\label{auto}
In the context of varying levels of uncertainty across environments, the selection of an appropriate expectile $\alpha$ becomes contingent on the specific characteristics of each environment. To automate the process of choosing the optimal expectile, we employ a bandit algorithm, specifically the Exponentially Weighted Average Forecasting algorithm \citep{cesa2006prediction}. We denote this method as \texttt{AutoExpectRL.} This formulation adopts the multi-armed bandit problem, where each bandit arm corresponds to a distinct value of $\alpha$. We consider a set of $D$ expectiles making predictions from a discrete set of values $\{\alpha_d \}_{d=1}^D$. At each episode $m$, a cumulative reward $R_m $ is sampled, and a distribution over arms $\mathbf{p}_m \in \Delta_D$ is formed, where $\mathbf{p}_m(d) \propto \exp \left(w_m(d)\right)$. The feedback signal $f_m \in \mathbb{R}$ is determined based on the arm selection as the improvement in performance, specifically $f_m=R_m-R_{m-1}$, where $R_m$ denotes the cumulative reward obtained in the episode $m$. Then, $w_{m+1}$ is obtained from $w_{m}$ by modifying only the $d_m$  according to $w_{m+1}(d_m) = w_m(d_m) + \eta \frac{f_m}{\mathbf{p}_m(d)} $
where $\eta>0$ is a step size parameter. 
The exponential weights distribution over $\alpha$ values at episode $m$ is denoted as $\mathbf{p}_m^\alpha$. This approach can be seen as a form of model selection akin to the methodology presented by \citet{pacchiano2020model}. Notably, instead of training distinct critics and actors for each $\alpha$ choice, our approach updates one single neural network for the critic and one single neural network for the actor. \looseness=-1
In both critic and actor, neural networks are composed of one common body and different heads for every value of $\alpha$, in our case $4$ values for $\left\{\alpha_d\right\}_{d=1}^D=\{0.2, 0.3, 0.4, 0.5\}$. The critic's heads correspond to the $4$ expectile losses for different values of $\alpha$. The actor's neural network is trained using $4$ classical TD3 losses, evaluated with action chosen by one specific head of the actor. Then in both critic and actor, the $4$ losses are summed, allowing an update of all heads at each iteration. Finally, the sampling of new trajectories is done using the chosen head of the actor, proposed by the bandit algorithm. More details about implementation can be found in Appendix \ref{appendix:auto}. Intuitively, when the agent receives a higher reward compared to the previous trajectory, the probability of choosing this arm is increased to encourage this arm to be picked again. Note that the use of a bandit algorithm to automatically select hyperparameters in an RL algorithm has been proposed in other contexts, such as  \citet{moskovitz2021tactical,badia2020agent57}. The \texttt{AutoExpectRL} method allows picking automatically expectile $\alpha$ and reduces hyperparameter tuning.  Practical details can be found in Appendix \ref{appendix:auto} where we expose the neural network architecture of this problem and associated losses. Note that this approach does not work in the DR setting as uncertainty parameters change between trajectories in DR. It is difficult for the algorithm to know if high or low rewards on trajectories come because the uncertainty parameter leads to small rewards, or if it is due to bad expectile picked at this iteration. 
\section{Empirical Result on Mujoco }
\begin{table*}[t]
\npdecimalsign{.}
\nprounddigits{2}
	\begin{center}
 
\tiny
\begin{tabular}
{|l|r|r|r|r|r|} \hline \ & TD3 Twin Critic& TD3 1 critic & \texttt{ExpectRL} best Expectile  & \texttt{AutoExpectRL} \\  \hline \hline $Ant (\times10^3)$  & $3.65  \pm 0.33$ & $1.90  \pm 0.07$ & $\mathbf{4.46 \pm 0.12}$  & $4.27 \pm 0.25$ \\ 
\hline $HalfCheetah (\times10^3)$ & $\mathbf{10.91}  \pm 0.14$ & $10.36  \pm 0.54$ & $10.42  \pm 0.13$ &$10.40 \pm 0.09$ \\ \hline $Hopper (\times10^3)$ & $2.88  \pm 0.10$ & $2.022  \pm 0.09$ & $\mathbf{3.10 \pm 0.05}$  & $3.03\pm 0.11$\\ \hline $Walker (\times10^3)$ & $2.95  \pm 0.12$ & $2.35  \pm 0.25$ & $\mathbf{3.22  \pm 0.11}$ & $3.02\pm 0.09 $\\  \hline $HumanoidStandup  (\times10^5)$ & $1.101  \pm 0.09$ & $1.087  \pm 0.09$ & $\mathbf{1.197  \pm 0.05}$  & $1.143\pm0.010 $\\ \hline \end{tabular}
\end{center}
\caption{Expectile vs Twin-critic, Mean performance $\pm$ standard error, on $10$ train seed }
\label{nominal}
\end{table*}
The Mujoco benchmark is employed in this experiment due to its significance for evaluating robustness in the context of continuous environments, where physical parameters may vary. In contrast, the Atari benchmark very deterministic with discrete action space without physical parameters cannot change during the testing period.
In this section, we compare the performance of the TD3 algorithm using the twin critic method during learning, only one critic, and finally our method \texttt{ExpectRL}. The different values of $\alpha$ are $\left\{\alpha_d\right\}_{d=1}^D=\{0.2, 0.3, 0.4 ,0.5    \}$. We can notice that  $\texttt{ExpectRL}$ with $\alpha=0.5$ is exactly TD3 with one critic.   Here, we only interact with the nominal and there is no notion of robustness. The mean and standard deviation are reported in Table \ref{nominal}, where we use $10$ seeds of $3$M steps for training, each evaluated on $30$ trajectories. The last column is our last algorithm, \texttt{AutoExpectRL}. \looseness=-1
In all environments except HalfCheetah, \texttt{ExpectRL} with fine-tuning of $\alpha$ has the best score and \texttt{AutoExpectRL} has generally close results. The scores for every expectiles can be found in Appendix\ref{appendix:score}.
In Halcheetah \ref{nominal} environment, it seems that no pessimism about $Q$-function is needed and our method \texttt{ExpectRL} is outperformed by TD3 with twin critic. Similar observations have been observed in \cite{moskovitz2021tactical} on this environment. Moreover, results for $\alpha=0.5$ and $\alpha=0.4$ are very close in Appendix~\ref{appendix:score} while the variance is reduced using $\alpha=0.4$. Results of Table \ref{nominal} show that it is possible to replace the twin critic approach with only one critic with the relevant value of pessimism or expectile. Moreover, one can remark in Appendix~\ref{appendix:score} that in  Hopper, Walker, and Ant environment, high pessimism is needed to get an accurate $Q$ function and better results, with a value of $\alpha=0.2$ or  $\alpha=0.3$ whereas less pessimism with $\alpha=0.4$ is needed for HumanoidStandup and HalfCheetah. Note that the value of $\alpha=0.5$ is never chosen and leads to generally the worst performance as reported in column TD3 with one critic which coincides with $\alpha=0.5$. Finally, the variance is also decreased using our method compared to TD3 with twin critics or TD3 with one critic. Finally, our method \texttt{AutoExectRL} allows choosing automatically the expectile almost without loss of performance and outperforming TD3, except on the environment HalfCheetah. Learning curves can be found in Appendix \ref{appendix:score}.\looseness=-1
\begin{table}[]
\begin{center}
\tiny
\begin{tabular}{|l|l|l|l|l|l|l|} \hline  & TD3 mean & \texttt{ExpectRL} mean & \texttt{Auto} mean& TD3 worst & \texttt{ExpectRL} worst  & \texttt{Auto} worst \\  \hline \hline $Ant1$ & $2.76  \pm 0.5$ & $3.55  \pm 0.65$& $\mathbf{3.55 \pm 0.51}$    & $2.22  \pm 0.5$ & $2.65  \pm 0.57$ & $ \mathbf{2.71\pm0.43} $ \\ 
\hline $Ant2$ & $2.28  \pm 0.09$ & $ \mathbf{2.50  \pm 0.89}$ & $2.41 \pm 0.77  $& $1.59  \pm 0.08$ & $\mathbf{2.49  \pm 0.94}$  & $2.42 \pm 0.51$\\
\hline $Ant3 $ & $0.31  \pm 1.13$ & $ \mathbf{0.54  \pm 0.08}$&  $0.53\pm 0.69$& $-0.99  \pm 1.13$ & $-0.94  \pm 0.21$ &  $\mathbf{-0.88 \pm0.34}$\\ 
\hline $Half1 $ & $2.79  \pm 0.22$ & $\mathbf{3.05  \pm 0.48}$ & $2.98  \pm 0.19$& $-0.34  \pm 0.04$ & $\mathbf{-0.27  \pm 0.19}$ &$-0.27  \pm 0.21$\\ 
\hline $Half2$ & $ \mathbf{2.63  \pm 0.20}$ & $2.51  \pm 0.41$ &   $2.58 \pm 0.32$& $-0.53  \pm 0.06$ & $\mathbf{-0.223  \pm 0.16}$ &  $-0.23  \pm 0.10$   \\ 
\hline $Half3 $ & $\mathbf{2.47  \pm 0.18}$ & $2.45  \pm 0.42$ & $2.39  \pm 0.15$  &$-0.61  \pm 0.08$ &  $\mathbf{-0.557  \pm 0.27}$ & $-0.58  \pm 0.09$  \\ 
\hline $Hopper 1 $ & $2.39  \pm 0.14$ & $\mathbf{2.76  \pm 0.04}$ &$2.52  \pm 0.11$ &$0.4  \pm 0.02$ & $0.44  \pm 0.01$  & $\mathbf{0.449  \pm 0.15}$ \\ 
\hline $Hopper 2 $ & $1.54  \pm 0.17$ & $ \mathbf{2.06  \pm 0.01}$ & $1.87  \pm 0.02$& $0.21  \pm 0.04$ & $\mathbf{0.32  \pm 0.03}$  &$0.32  \pm 0.03$ 
\\ \hline $Hopper 3 $ & $1.15  \pm 0.14$ & $\mathbf{1.43  \pm 0.02}$ &$1.433  \pm 0.09$  & $0.14  \pm 0.03$ & $ \mathbf{0.25  \pm 0.22}$ & $2.42  \pm 0.19$ 
\\ \hline $Walker 1 $ & $3.12  \pm 0.2$ & $ \mathbf{3.66  \pm 0.68}$ &  $3.58  \pm 0.27$    &$0.68  \pm 0.12$ & $ \mathbf{2.77  \pm 0.15}$ & $1.99  \pm 0.13 $
\\ \hline $Walker 2$ & $2.70  \pm 0.2$ & $\mathbf{3.98}  \pm 0.58$ & $3.88  \pm 0.61$ &$0.28  \pm 0.07$ & $ \mathbf{1.36  \pm 0.82}$  & $1.11  \pm 0.15$ 
\\ \hline $Walker 3 $ & $2.60  \pm 0.18$ & $\mathbf{3.84  \pm 0.45}$ &$3.58  \pm 0.15$ &$0.17  \pm 0.06$ & $ 0.65  \pm 0.12$ &$ \mathbf{0.87  \pm 0.09}$
\\ \hline $Humanoid1 $ & $1.03  \pm 0.4$ & $1.12  \pm 0.25$ &  $\mathbf{1.13  \pm 0.26}$     &$0.85  \pm 0.07$ & $\mathbf{0.97  \pm 0.23}$ & $0.98  \pm 0.24$\\ \hline $Humanoid2 $ & $1.03  \pm 0.3$ & $\mathbf{1.13  \pm 0.15}$ & $1.11  \pm 0.12$&$0.73  \pm 0.07$ & $ \mathbf{0.83  \pm 0.23}$& $0.80  \pm 0.18$ \\  \hline $Humanoid3 $ & $1.01  \pm 0.3$ & $ \mathbf{1.06  \pm 0.13}$ & $1.05  \pm 0.18$ &$0.57  \pm 0.04$ & $ \mathbf{0.71  \pm 0.21}$&  $0.68  \pm 0.09$
\\ \hline
\end{tabular}
\caption{Result on Robust Benchmark for TD3  \texttt{ExpectRL} and \texttt{AutoExpectRL}. Results are $\times10^3$ bigger for all environments except for Humanoid where results are $\times10^5$ bigger.}
\label{classical}
\end{center}
\end{table}
\section{Empirical Results on Robust Benchmark }

\begin{table}
\tiny
\begin{center}
\begin{tabular}{|l|l|l|l|l|l|l|} \hline &DR+\texttt{ExpectRL}(m)& M2TD3(m)& DR(m)&DR+\texttt{ExpectRL}(w)& M2TD3(w)& DR(w)\\  \hline \hline$Ant1 $ & $4.84  \pm 0.43$ & $4.51  \pm 0.08$ & $\mathbf{5.25  \pm 0.1}$ & $3.36  \pm 0.55$ & $\mathbf{3.84  \pm 0.1}$ & $3.51  \pm 0.08$ \\\hline $Ant2 $ & $5.63  \pm 0.43$ & $5.44  \pm 0.05$ & $\mathbf{6.32  \pm 0.09}$ & $2.72  \pm 0.42$ & $\mathbf{4.13  \pm 0.11}$ & $1.64  \pm 0.13$ \\ \hline$Ant3 $ & $2.86  \pm 1.03$ & $2.66  \pm 0.22$ & $\mathbf{3.62  \pm 0.11}$ & $\mathbf{0.28  \pm 0.35}$ & $0.10  \pm 0.10$ & $-0.32  \pm 0.03$ \\\hline $Half 1 $ & $5.3  \pm 0.59$ & $3.89  \pm 0.06$ & $\mathbf{5.93  \pm 0.18}$ & $2.86  \pm 0.99$ & $3.14  \pm 0.10$ & $\mathbf{3.19  \pm 0.08}$ \\ \hline$Half2 $ & $5.25  \pm 0.32$ & $4.35  \pm 0.05$ & $\mathbf{5.79  \pm 0.15}$ & $1.77  \pm 0.31$ & $\mathbf{2.61  \pm 0.16}$ & $2.12  \pm 0.13$ \\ \hline$Half3 $ & $4.52  \pm 0.24$ & $3.79  \pm 0.09$ & $\mathbf{5.54  \pm 0.16}$ & $1.02  \pm 0.24$ & $0.93  \pm 0.21$ & $\mathbf{1.09  \pm 0.06}$ \\\hline $Hopper 1 $ & $2.58  \pm 0.23$ & $\mathbf{2.68  \pm 0.11}$ & $2.57  \pm 0.15$ & $\mathbf{0.64  \pm 0.20}$ & $0.62  \pm 0.45$ & $0.53  \pm 0.26$ \\ \hline$Hopper 2 $ & $\mathbf{2.53  \pm 0.22}$ & $2.51  \pm 0.07$ & $1.89  \pm 0.08$ & $\mathbf{0.55  \pm 0.07}$ & $0.53  \pm 0.28$ & $0.47  \pm 0.02$ \\ \hline$Hopper 3 $ & $\mathbf{2.21  \pm 0.33}$ & $0.85  \pm 0.07$ & $1.5  \pm 0.07$ & $\mathbf{0.39  \pm 0.07}$ & $0.28  \pm 0.25$ & $0.21  \pm 0.03$ \\ \hline $Walker 1 $ & $\mathbf{3.77  \pm 0.89}$ & $3.70  \pm 0.31$ & $3.59  \pm 0.26$ & $\mathbf{3.41  \pm 0.05}$ & $2.83  \pm 0.39$ & $2.19  \pm 0.42$ \\ \hline$Walker 2 $ & $\mathbf{4.75  \pm 0.57}$ & $4.72  \pm 0.12$ & $4.54  \pm 0.31$ & $2.74  \pm 0.61$ & $\mathbf{3.14  \pm 0.39}$ & $2.31  \pm 0.51$ \\ \hline$Walker 3$ & $4.39  \pm 0.37$ & $4.27  \pm 0.21$ & $\mathbf{4.48  \pm 0.16}$ & $1.14  \pm 0.79$ & $\mathbf{1.34  \pm 0.43}$ & $1.32  \pm 0.34$ \\  \hline $Humanoid1 $ & $\mathbf{1.21  \pm 0.23}$ & $1.08  \pm 0.04$ & $1.12  \pm 0.05$ & $\mathbf{1.04  \pm 0.86}$ & $0.93  \pm 0.07$ & $0.96  \pm 0.06$ \\ \hline$Humanoid 2 $ & $\mathbf{1.23  \pm 0.22}$ & $0.97  \pm 0.04$ & $1.06  \pm 0.04$ & $\mathbf{0.86  \pm 0.28}$ & $0.65  \pm 0.07$ & $0.73  \pm 0.78$ \\ \hline $Humanoid 3 $ & $\mathbf{1.12  \pm 0.35}$ & $1.09  \pm 0.06$ & $1.04  \pm 0.07$ & $\mathbf{0.84  \pm 0.26}$ & $0.62  \pm 0.06$ & $0.54  \pm 0.34$  \\\hline
\end{tabular}
\end{center}
\caption{Result on Robust Benchmark  for \texttt{ExpectRL} + DR , M2TD3 and DR. Results are $\times10^3$ bigger for all environments except for Humanoid results are $\times10^5$ bigger.  The mean performance is denoted $(m)$ and worst case $(w)$.}
\label{benchmark:robust}
\end{table}
This section presents an assessment of the worst-case and average performance and generalization capabilities of the proposed algorithm. The experimental validation was conducted on optimal control problems utilizing the MuJoCo simulation environments \citep{todorov2012mujoco}.
The performance of the algorithm was systematically benchmarked against state-of-the-art robust RL M2TD3  as it is state of the art compared to other algorithms methodologies, 
 M3DDPG, and RARL. Furthermore, a comparative analysis was undertaken with Domain Randomization (DR) as introduced by \citet{tobin2017domain} for a comprehensive evaluation. \looseness=-1
To assess the worst-case performance of the policy  $\pi$ under varying uncertainty parameters $\omega \in \Omega$, following the benchmark of \citet{tanabe2022max}, $30$ evaluations of the cumulative reward were conducted for each uncertainty parameter value $\omega_1, \ldots, \omega_K \in \Omega$. Specifically, $R_k(\pi)$ denotes the cumulative reward on $\omega_k$, averaged over $30$ trials. Subsequently, $R_{\text{worst}}(\pi) = \min_{1 \leqslant k \leqslant K} R_k(\pi)$ (denoted (w) in Table \ref{classical} and \ref{benchmark:robust}) was computed as an estimate of the worst-case performance of $\pi$ on $\Omega$. Additionally, the average performance was computed as $R_{\text{average}}(\pi) = \frac{1}{K} \sum_{k=1}^K R_k(\pi)$ (denoted (m) in Table \ref{classical} and \ref{benchmark:robust}). For the evaluation process, $K$ uncertainty parameters $\omega_1, \ldots, \omega_K$ were chosen according to the dimensionality of $\omega$: for $1$D $\omega$, $K=10$ equally spaced points on the $1$D interval $\Omega$; for $2$D $\omega$, $10$ equally spaced points were chosen in each dimension of $\Omega$, resulting in $K=100$ points; and for $3$D $\omega$, $10$ equally spaced points were selected in each dimension of $\Omega$, resulting in $K=1000$ points or different environments. Each approach underwent policy training $10$ times in each environment. The training time steps $T_{\max}$ were configured as $2 \mathrm{M}, 4 \mathrm{M}$, and $5 \mathrm{M}$ for scenarios with $1 \mathrm{D}, 2 \mathrm{D}$, and $3 \mathrm{D}$ uncertainty parameters respectively,  following \citet{tanabe2022max}.  Table \ref{table:environment_settings} summarizes the different changes of parameters in the environments. The final policies obtained from training were then evaluated for their worst-case performances and average performance over all uncertainty parameters. The results are the following.

\vspace{0.3cm}

We first demonstrate that our method \texttt{ExpectRl} is more robust than the classical RL algorithm. To do so, we conduct the benchmark task presented previously on TD3 algorithm (with twin critic trick) as a baseline and our method \texttt{ExpectRl}. As exposed in Table \ref{classical}, our method outperforms TD3 in all environments on worst-case performance, which was expected as TD3 is not designed by nature to be robust and to maximize a worst-case performance. Moreover, \texttt{AutoExpectRL} has good and similar performance compared to the best expectile like in Table \ref{nominal}.
As TD3 has sometimes very bad performance, our method also performs better on average over all environments except HalfCheetah 2 and HalfCheetah 3.
These two environments required more exploration, and pessimism is in general not a good thing for these tasks. 
Moreover, robustness is not needed in HalfCheetah environments that are already quite stable compared to other tasks in Mujoco. However, \texttt{ExpectRL} needs to be compared with algorithms designed to be robust, such as M2TD3 which has state-of-the-art performance on this benchmark. \looseness=-1

\vspace{0.3cm}

If performance of \texttt{ExpectRL} in Table \ref{classical} and the performance of M2TD3 in Table \ref{benchmark:robust} are compared, 
we can observe a large difference on many tasks where M2TD3 outperforms, in general, our method. 
This is because sampling trajectories from the entire uncertainty set allows M2TD3 to get knowledge from dangerous trajectories and allows the algorithm to generalize better than our method, which only samples from the nominal. The comparison between methods is then not fair for \texttt{ExpectRl} which has only access to samples from the nominal and this is why the method \texttt{ExpectRL + DR} was introduced. Receiving information about all environments that need to be robust during the training phase, the algorithm tends to obtain better performance on minimum performance over these environments on testing. Table \ref{benchmark:robust} shows the result on average and on worst-case performance between our second method \texttt{ExpectRL} + DR with tuning of $\alpha$ against M2TD3 and DR approach. Recall that \texttt{AutoExpectRL} cannot be used with DR as mentioned at the end of Section \ref{auto}. \looseness=-1

\vspace{0.3cm}

In terms of worst-case performance, our method outperforms 9 times M2TD3 (8 times in bold and one time when DR is better in general for HalfCheetaht3) and has a worse performance on 6 tasks compared to M2TD3. Our method is therefore competitive with the state of the art in robust algorithms such as M2TD3, which already outperformed M3DDPG and RARL  on worst-case performance. Except on Hopper1, our method outperforms M2TD3 on average, results which show that M2TD3 is very pessimistic compared to our method. However, 
in terms of average results, we can see that DR, which is designed to be good on average across all environments, generally performs better than our method and M2TD3 expect on Hopper, Walker1 and 2, and HumanoidStandup which are not stable and need to be robustified to avoid catastrophic performance that affect too much the mean performance over all environment. Moreover, compared to M2TD3, our method \texttt{ExpecRL}, even without auto fine-tuning of $\alpha$, has the advantage of having fewer parameter tuning compared to the M2TD3 algorithm. \looseness=-1
\section{Conclusion and perspectives}
We propose a simple method, \texttt{ExpectRL} to replace twin critic in practice, only replacing the classic $L_2$ loss of the critic with an expectile loss. Moreover, we show that it can also lead to a Robust RL algorithm and demonstrate the effectiveness of our method combined with DR on a robust RL Benchmark. The limitations of our method are that \texttt{AutoExpectRL} allows fine-tuning of $\alpha$ only without combining with DR. Another limitation is that the uncertainty set defined with expectile is not very interpretable, and other algorithms with a more interpretable set and similar experimental results would be interesting. About future perspectives, we demonstrate the effectiveness of our method using as baselines TD3, but our method can be easily adapted to any algorithm using a $Q$-function such as classical DQN, SAC, and other algorithms both with discrete or continuous action space.
Finally, theoretically, it would be interesting to study for example sample complexity of this method compared to the classical RL algorithm. \looseness=-1

\label{submission}


\newpage

\bibliography{bib}

\begin{thebibliography}{53}
\providecommand{\natexlab}[1]{#1}
\providecommand{\url}[1]{\texttt{#1}}
\expandafter\ifx\csname urlstyle\endcsname\relax
  \providecommand{\doi}[1]{doi: #1}\else
  \providecommand{\doi}{doi: \begingroup \urlstyle{rm}\Url}\fi

\bibitem[Abdullah et~al.(2019)Abdullah, Ren, Ammar, Milenkovic, Luo, Zhang, and Wang]{abdullah2019wasserstein}
Abdullah, M.~A., Ren, H., Ammar, H.~B., Milenkovic, V., Luo, R., Zhang, M., and Wang, J.
\newblock Wasserstein robust reinforcement learning.
\newblock \emph{arXiv preprint arXiv:1907.13196}, 2019.

\bibitem[Artzner et~al.(1999)Artzner, Delbaen, Eber, and Heath]{artzner1999coherent}
Artzner, P., Delbaen, F., Eber, J.-M., and Heath, D.
\newblock Coherent measures of risk.
\newblock \emph{Mathematical finance}, 9\penalty0 (3):\penalty0 203--228, 1999.

\bibitem[Badia et~al.(2020)Badia, Piot, Kapturowski, Sprechmann, Vitvitskyi, Guo, and Blundell]{badia2020agent57}
Badia, A.~P., Piot, B., Kapturowski, S., Sprechmann, P., Vitvitskyi, A., Guo, Z.~D., and Blundell, C.
\newblock Agent57: Outperforming the atari human benchmark.
\newblock In \emph{International conference on machine learning}, pp.\  507--517. PMLR, 2020.

\bibitem[Bellini \& Di~Bernardino(2017)Bellini and Di~Bernardino]{bellini2017risk}
Bellini, F. and Di~Bernardino, E.
\newblock Risk management with expectiles.
\newblock \emph{The European Journal of Finance}, 23\penalty0 (6):\penalty0 487--506, 2017.

\bibitem[Bellini et~al.(2014)Bellini, Klar, M{\"u}ller, and Gianin]{bellini2014generalized}
Bellini, F., Klar, B., M{\"u}ller, A., and Gianin, E.~R.
\newblock Generalized quantiles as risk measures.
\newblock \emph{Insurance: Mathematics and Economics}, 54:\penalty0 41--48, 2014.

\bibitem[Brekelmans et~al.(2022)Brekelmans, Genewein, Grau-Moya, Del{\'e}tang, Kunesch, Legg, and Ortega]{brekelmans2022your}
Brekelmans, R., Genewein, T., Grau-Moya, J., Del{\'e}tang, G., Kunesch, M., Legg, S., and Ortega, P.
\newblock Your policy regularizer is secretly an adversary.
\newblock \emph{arXiv preprint arXiv:2203.12592}, 2022.

\bibitem[Cesa-Bianchi \& Lugosi(2006)Cesa-Bianchi and Lugosi]{cesa2006prediction}
Cesa-Bianchi, N. and Lugosi, G.
\newblock \emph{Prediction, learning, and games}.
\newblock Cambridge university press, 2006.

\bibitem[Chow et~al.(2015)Chow, Tamar, Mannor, and Pavone]{chow2015risk}
Chow, Y., Tamar, A., Mannor, S., and Pavone, M.
\newblock Risk-sensitive and robust decision-making: a cvar optimization approach.
\newblock \emph{Advances in neural information processing systems}, 28, 2015.

\bibitem[Clavier et~al.(2022)Clavier, Allassoni{\`e}re, and Pennec]{clavier2022robust}
Clavier, P., Allassoni{\`e}re, S., and Pennec, E.~L.
\newblock Robust reinforcement learning with distributional risk-averse formulation.
\newblock \emph{arXiv preprint arXiv:2206.06841}, 2022.

\bibitem[Clavier et~al.(2023)Clavier, Pennec, and Geist]{clavier2023towards}
Clavier, P., Pennec, E.~L., and Geist, M.
\newblock Towards minimax optimality of model-based robust reinforcement learning.
\newblock \emph{arXiv preprint arXiv:2302.05372}, 2023.

\bibitem[Dabney et~al.(2018)Dabney, Ostrovski, Silver, and Munos]{dabney2018implicit}
Dabney, W., Ostrovski, G., Silver, D., and Munos, R.
\newblock Implicit quantile networks for distributional reinforcement learning.
\newblock In \emph{International conference on machine learning}, pp.\  1096--1105. PMLR, 2018.

\bibitem[Delbaen(2000)]{delbaen2000draft}
Delbaen, F.
\newblock Draft: Coherent risk measures.
\newblock \emph{Lecture notes, Pisa}, 2000.

\bibitem[Delbaen(2002)]{delbaen2002coherent}
Delbaen, F.
\newblock Coherent risk measures on general probability spaces.
\newblock \emph{Advances in finance and stochastics: essays in honour of Dieter Sondermann}, pp.\  1--37, 2002.

\bibitem[Delbaen(2013)]{delbaen2013remark}
Delbaen, F.
\newblock A remark on the structure of expectiles.
\newblock \emph{arXiv preprint arXiv:1307.5881}, 2013.

\bibitem[Derman \& Mannor(2020)Derman and Mannor]{derman2020distributional}
Derman, E. and Mannor, S.
\newblock Distributional robustness and regularization in reinforcement learning.
\newblock \emph{arXiv preprint arXiv:2003.02894}, 2020.

\bibitem[Derman et~al.(2021)Derman, Geist, and Mannor]{derman2021twice}
Derman, E., Geist, M., and Mannor, S.
\newblock Twice regularized mdps and the equivalence between robustness and regularization.
\newblock \emph{Advances in Neural Information Processing Systems}, 34, 2021.

\bibitem[Eysenbach \& Levine(2021)Eysenbach and Levine]{eysenbach2021maximum}
Eysenbach, B. and Levine, S.
\newblock Maximum entropy rl (provably) solves some robust rl problems.
\newblock \emph{arXiv preprint arXiv:2103.06257}, 2021.

\bibitem[Fujimoto et~al.(2018)Fujimoto, Hoof, and Meger]{fujimoto2018addressing}
Fujimoto, S., Hoof, H., and Meger, D.
\newblock Addressing function approximation error in actor-critic methods.
\newblock In \emph{International conference on machine learning}, pp.\  1587--1596. PMLR, 2018.

\bibitem[Gneiting(2011)]{gneiting2011making}
Gneiting, T.
\newblock Making and evaluating point forecasts.
\newblock \emph{Journal of the American Statistical Association}, 106\penalty0 (494):\penalty0 746--762, 2011.

\bibitem[Greenberg et~al.(2022)Greenberg, Chow, Ghavamzadeh, and Mannor]{greenberg2022efficient}
Greenberg, I., Chow, Y., Ghavamzadeh, M., and Mannor, S.
\newblock Efficient risk-averse reinforcement learning.
\newblock \emph{Advances in Neural Information Processing Systems}, 35:\penalty0 32639--32652, 2022.

\bibitem[Haarnoja et~al.(2018)Haarnoja, Zhou, Abbeel, and Levine]{haarnoja2018soft}
Haarnoja, T., Zhou, A., Abbeel, P., and Levine, S.
\newblock Soft actor-critic: Off-policy maximum entropy deep reinforcement learning with a stochastic actor.
\newblock In \emph{International conference on machine learning}, pp.\  1861--1870. PMLR, 2018.

\bibitem[Hasselt(2010)]{hasselt2010double}
Hasselt, H.
\newblock Double q-learning.
\newblock \emph{Advances in neural information processing systems}, 23, 2010.

\bibitem[Ho et~al.(2021)Ho, Petrik, and Wiesemann]{ho2021partial}
Ho, C.~P., Petrik, M., and Wiesemann, W.
\newblock Partial policy iteration for l1-robust markov decision processes.
\newblock \emph{J. Mach. Learn. Res.}, 22:\penalty0 275--1, 2021.

\bibitem[Husain et~al.(2021)Husain, Ciosek, and Tomioka]{husain2021regularized}
Husain, H., Ciosek, K., and Tomioka, R.
\newblock Regularized policies are reward robust.
\newblock In \emph{International Conference on Artificial Intelligence and Statistics}, pp.\  64--72. PMLR, 2021.

\bibitem[Iyengar(2005)]{iyengar2005robust}
Iyengar, G.~N.
\newblock Robust dynamic programming.
\newblock \emph{Mathematics of Operations Research}, 30\penalty0 (2):\penalty0 257--280, 2005.

\bibitem[Jullien et~al.(2023)Jullien, Deffayet, Renders, Groth, and de~Rijke]{jullien2023distributional}
Jullien, S., Deffayet, R., Renders, J.-M., Groth, P., and de~Rijke, M.
\newblock Distributional reinforcement learning with dual expectile-quantile regression.
\newblock \emph{arXiv preprint arXiv:2305.16877}, 2023.

\bibitem[Kostrikov et~al.(2021)Kostrikov, Nair, and Levine]{kostrikov2021offline}
Kostrikov, I., Nair, A., and Levine, S.
\newblock Offline reinforcement learning with implicit q-learning.
\newblock \emph{arXiv preprint arXiv:2110.06169}, 2021.

\bibitem[Kumar et~al.(2022)Kumar, Levy, Wang, and Mannor]{kumar2022efficient}
Kumar, N., Levy, K., Wang, K., and Mannor, S.
\newblock Efficient policy iteration for robust markov decision processes via regularization.
\newblock \emph{arXiv preprint arXiv:2205.14327}, 2022.

\bibitem[Kumar et~al.(2023)Kumar, Derman, Geist, Levy, and Mannor]{kumar2023policy}
Kumar, N., Derman, E., Geist, M., Levy, K., and Mannor, S.
\newblock Policy gradient for s-rectangular robust markov decision processes.
\newblock \emph{arXiv preprint arXiv:2301.13589}, 2023.

\bibitem[Li et~al.(2023)Li, Sutter, and Kuhn]{li2023policy}
Li, M., Sutter, T., and Kuhn, D.
\newblock Policy gradient algorithms for robust mdps with non-rectangular uncertainty sets.
\newblock \emph{arXiv preprint arXiv:2305.19004}, 2023.

\bibitem[Li et~al.(2019)Li, Wu, Cui, Dong, Fang, and Russell]{li2019robust}
Li, S., Wu, Y., Cui, X., Dong, H., Fang, F., and Russell, S.
\newblock Robust multi-agent reinforcement learning via minimax deep deterministic policy gradient.
\newblock In \emph{Proceedings of the AAAI conference on artificial intelligence}, volume~33, pp.\  4213--4220, 2019.

\bibitem[Lillicrap et~al.(2015)Lillicrap, Hunt, Pritzel, Heess, Erez, Tassa, Silver, and Wierstra]{lillicrap2015continuous}
Lillicrap, T.~P., Hunt, J.~J., Pritzel, A., Heess, N., Erez, T., Tassa, Y., Silver, D., and Wierstra, D.
\newblock Continuous control with deep reinforcement learning.
\newblock \emph{arXiv preprint arXiv:1509.02971}, 2015.

\bibitem[Moos et~al.(2022)Moos, Hansel, Abdulsamad, Stark, Clever, and Peters]{moos2022robust}
Moos, J., Hansel, K., Abdulsamad, H., Stark, S., Clever, D., and Peters, J.
\newblock Robust reinforcement learning: A review of foundations and recent advances.
\newblock \emph{Machine Learning and Knowledge Extraction}, 4\penalty0 (1):\penalty0 276--315, 2022.

\bibitem[Moskovitz et~al.(2021)Moskovitz, Parker-Holder, Pacchiano, Arbel, and Jordan]{moskovitz2021tactical}
Moskovitz, T., Parker-Holder, J., Pacchiano, A., Arbel, M., and Jordan, M.
\newblock Tactical optimism and pessimism for deep reinforcement learning.
\newblock \emph{Advances in Neural Information Processing Systems}, 34:\penalty0 12849--12863, 2021.

\bibitem[Nilim \& El~Ghaoui(2005)Nilim and El~Ghaoui]{nilim2005robust}
Nilim, A. and El~Ghaoui, L.
\newblock Robust control of markov decision processes with uncertain transition matrices.
\newblock \emph{Operations Research}, 53\penalty0 (5):\penalty0 780--798, 2005.

\bibitem[Pacchiano et~al.(2020)Pacchiano, Phan, Abbasi~Yadkori, Rao, Zimmert, Lattimore, and Szepesvari]{pacchiano2020model}
Pacchiano, A., Phan, M., Abbasi~Yadkori, Y., Rao, A., Zimmert, J., Lattimore, T., and Szepesvari, C.
\newblock Model selection in contextual stochastic bandit problems.
\newblock \emph{Advances in Neural Information Processing Systems}, 33:\penalty0 10328--10337, 2020.

\bibitem[Pan et~al.(2019)Pan, Seita, Gao, and Canny]{pan2019risk}
Pan, X., Seita, D., Gao, Y., and Canny, J.
\newblock Risk averse robust adversarial reinforcement learning.
\newblock In \emph{2019 International Conference on Robotics and Automation (ICRA)}, pp.\  8522--8528. IEEE, 2019.

\bibitem[Pinto et~al.(2017)Pinto, Davidson, Sukthankar, and Gupta]{pinto2017robust}
Pinto, L., Davidson, J., Sukthankar, R., and Gupta, A.
\newblock Robust adversarial reinforcement learning.
\newblock In \emph{International Conference on Machine Learning}, pp.\  2817--2826. PMLR, 2017.

\bibitem[Raffin et~al.(2021)Raffin, Hill, Gleave, Kanervisto, Ernestus, and Dormann]{stable-baselines3}
Raffin, A., Hill, A., Gleave, A., Kanervisto, A., Ernestus, M., and Dormann, N.
\newblock Stable-baselines3: Reliable reinforcement learning implementations.
\newblock \emph{Journal of Machine Learning Research}, 22\penalty0 (268):\penalty0 1--8, 2021.
\newblock URL \url{http://jmlr.org/papers/v22/20-1364.html}.

\bibitem[Rowland et~al.(2019)Rowland, Dadashi, Kumar, Munos, Bellemare, and Dabney]{rowland2019statistics}
Rowland, M., Dadashi, R., Kumar, S., Munos, R., Bellemare, M.~G., and Dabney, W.
\newblock Statistics and samples in distributional reinforcement learning.
\newblock In \emph{International Conference on Machine Learning}, pp.\  5528--5536. PMLR, 2019.

\bibitem[Shi \& Chi(2022)Shi and Chi]{shi2022distributionally}
Shi, L. and Chi, Y.
\newblock Distributionally robust model-based offline reinforcement learning with near-optimal sample complexity.
\newblock \emph{arXiv preprint arXiv:2208.05767}, 2022.

\bibitem[Shi et~al.(2023)Shi, Li, Wei, Chen, Geist, and Chi]{shi2023curious}
Shi, L., Li, G., Wei, Y., Chen, Y., Geist, M., and Chi, Y.
\newblock The curious price of distributional robustness in reinforcement learning with a generative model.
\newblock \emph{arXiv preprint arXiv:2305.16589}, 2023.

\bibitem[Tanabe et~al.(2022)Tanabe, Sato, Fukuchi, Sakuma, and Akimoto]{tanabe2022max}
Tanabe, T., Sato, R., Fukuchi, K., Sakuma, J., and Akimoto, Y.
\newblock Max-min off-policy actor-critic method focusing on worst-case robustness to model misspecification.
\newblock \emph{Advances in Neural Information Processing Systems}, 35:\penalty0 6967--6981, 2022.

\bibitem[Tessler et~al.(2019)Tessler, Efroni, and Mannor]{tessler2019action}
Tessler, C., Efroni, Y., and Mannor, S.
\newblock Action robust reinforcement learning and applications in continuous control.
\newblock In \emph{International Conference on Machine Learning}, pp.\  6215--6224. PMLR, 2019.

\bibitem[Tobin et~al.(2017)Tobin, Fong, Ray, Schneider, Zaremba, and Abbeel]{tobin2017domain}
Tobin, J., Fong, R., Ray, A., Schneider, J., Zaremba, W., and Abbeel, P.
\newblock Domain randomization for transferring deep neural networks from simulation to the real world.
\newblock In \emph{2017 IEEE/RSJ international conference on intelligent robots and systems (IROS)}, pp.\  23--30. IEEE, 2017.

\bibitem[Todorov et~al.(2012)Todorov, Erez, and Tassa]{todorov2012mujoco}
Todorov, E., Erez, T., and Tassa, Y.
\newblock Mujoco: A physics engine for model-based control.
\newblock In \emph{2012 IEEE/RSJ international conference on intelligent robots and systems}, pp.\  5026--5033. IEEE, 2012.

\bibitem[Van~Hasselt et~al.(2016)Van~Hasselt, Guez, and Silver]{van2016deep}
Van~Hasselt, H., Guez, A., and Silver, D.
\newblock Deep reinforcement learning with double q-learning.
\newblock In \emph{Proceedings of the AAAI conference on artificial intelligence}, volume~30, 2016.

\bibitem[Wang et~al.(2023)Wang, Gadot, Kumar, Levy, and Mannor]{wang2023robust}
Wang, K., Gadot, U., Kumar, N., Levy, K., and Mannor, S.
\newblock Robust reinforcement learning via adversarial kernel approximation.
\newblock \emph{arXiv preprint arXiv:2306.05859}, 2023.

\bibitem[Yang et~al.(2021)Yang, Zhang, and Zhang]{yang2021towards}
Yang, W., Zhang, L., and Zhang, Z.
\newblock Towards theoretical understandings of robust markov decision processes: Sample complexity and asymptotics.
\newblock \emph{arXiv preprint arXiv:2105.03863}, 2021.

\bibitem[Ying et~al.(2021)Ying, Zhou, Su, Yan, and Zhu]{ying2021towards}
Ying, C., Zhou, X., Su, H., Yan, D., and Zhu, J.
\newblock Towards safe reinforcement learning via constraining conditional value-at-risk.
\newblock 2021.

\bibitem[Zhang et~al.(2023)Zhang, Hu, and Li]{zhang2023regularized}
Zhang, R., Hu, Y., and Li, N.
\newblock Regularized robust mdps and risk-sensitive mdps: Equivalence, policy gradient, and sample complexity.
\newblock \emph{arXiv preprint arXiv:2306.11626}, 2023.

\bibitem[Zhang et~al.(2021)Zhang, Liu, and Whiteson]{zhang2021mean}
Zhang, S., Liu, B., and Whiteson, S.
\newblock Mean-variance policy iteration for risk-averse reinforcement learning.
\newblock In \emph{Proceedings of the AAAI Conference on Artificial Intelligence}, volume~35, pp.\  10905--10913, 2021.

\bibitem[Zouitine et~al.(2023)Zouitine, Rachelson, and Geist]{zouitine2023revisiting}
Zouitine, A., Rachelson, E., and Geist, M.
\newblock Revisiting the static model in robust reinforcement learning.
\newblock In \emph{Sixteenth European Workshop on Reinforcement Learning}, 2023.

\end{thebibliography}
\bibliographystyle{icml2024}

\appendix
\section{Proof }
\label{contraction}
\begin{theorem}

    \begin{equation}
    (T^\pi_\alpha v)(s) = \sum_a \pi(a|s) (r(s,a) + \gamma m_\alpha(P_{sa}, v)).
\end{equation}
this is a contraction:
\end{theorem}

The expectile satisfies the following properties \citep{bellini2014generalized,bellini2017risk}:
\begin{enumerate}
    \item Translation invariance:  $ m_\tau(X+h) = m_\alpha(X) + h$
    
    \item Monotonicity:  $ X\leq Y \text{ a.s.} \Rightarrow m_\tau(X) \leq m_\alpha(Y)$  
    
    \item Positive homogeneity: 
 
 $\lambda\geq 0 \Rightarrow m_\alpha(\lambda X) = \lambda m_\tau(X)$
    
    \item Superadditivity, for $\alpha\leq\frac{1}{2}, m_\tau(X+Y) \geq m_\alpha(X) + m_\alpha(Y).$
\end{enumerate}
So,
\begin{align}
   & (T^\pi_\alpha v_1)(s) - (T^\pi_\alpha v_2)(s) 
    = \sum_a \pi(a|s) (r(s,a) + \gamma m_\alpha(P_{sa}, v_1)) \\&- \sum_a \pi(a|s) (r(s,a) + \gamma m_\alpha(P_{sa}, v_2)) 
    \\
    &= \gamma \sum_a \pi(a|s) (m_\alpha(P_{sa}, v_1) - m_\alpha(P_{sa}, v_2) )
    \\
    &\leq \gamma \sum_a \pi(a|s) (m_\alpha(P_{sa}, v_2 + \|v_2-v_1\|_\infty) - m_\alpha(P_{sa}, v_2) ) \\
    & \notag \text{ (by monotonicity) and $v_1\leq v_2 + \|v_2-v_1\|_\infty$)}\\
    &= \gamma \sum_a  \pi(a|s) (m_\alpha(P_{sa}, v_2) + \|v_2-v_1\|_\infty - m_\alpha(P_{sa}, v_2)) \text{ (by translation invariance)}
    \\
    &= \gamma \|v_2 - v_1\|_\infty.
\end{align}

In the same manner, $T_\alpha^*$ is also a contraction, as the only line of this proof that differs is replacing the expectation by a $\max _a$. As maximum operator $1$-Lipschitz, (ie)  $\max_a  f(a) -\max g(a)\leq \max f(a)-g(a) $, we obtain $\gamma$- contraction results also for the optimal Bellman operator $T_\alpha^*$.

Similar ideas exist in \citet{zhang2023regularized}, which show similar properties for risk-sensitive MDPs defined through a convex risk measure, even though they do not consider explicitly the expectile which is a convex risk measure for $\alpha<1/2$.

\begin{theorem} \label{appendix:link_robust_sensitive}
    The (optimal) Expectile value function is equal to the (optimal) robust value function
    \begin{align}
         &v^*_\alpha(s)=v^\pi_{\mathcal{E}}:=\max_\pi \min _{Q \in \mathcal{E}} v^{\pi,Q}   \\
&v^\pi_\alpha(s)=v^\pi_{\mathcal{E}}:= \min _{Q \in \mathcal{E}} v^{\pi,Q}   
    \end{align}
    
\end{theorem}
where $\mathcal{E}$ is defined in section \ref{exextile} or below. 
\begin{proof}
   This theorem is just an adaptation of  Theorem 2 in \cite{zhang2023regularized} where we use expectile risk measure $m_\alpha(X)$ which  implicitly defined the uncertainty set for robust $\mathcal{E}$ such that :
\begin{align*}
&m_\alpha(X)=\min _{Q \in \mathcal{E}}\mathbb{E}_{Q}[X] ;\\
     &\mathcal{E}=\left\{ Q \in \mathcal{P} \mid\  \exists \eta>0,   \sqrt{\frac{\alpha}{1-\alpha}} \eta \leq \frac{d Q}{d P} \leq \sqrt{\frac{(1-\alpha)}{\alpha}} \eta \right\} 
\end{align*}
where $\mathcal{P}$ is the set of $P$-absolutely continuous probability measures.
In  Theorem \cite{zhang2023regularized},  they link Risk sensitive MDPs (in our case expectile formulation) with Regularised Robust MDPs. In our case, we can rewrite the classical RMDPs to Regularised-Robust MDPs such that:

\begin{align*}
    v^*_\mathcal{E}&= \max _\pi \min _{Q \in \mathcal{E}} v^{\pi,Q} =  \max _\pi \min _{Q \in \mathcal{E}} \mathbb{E}\Big[  \sum_t \gamma^t r(s_t,a_t)    \Big]    \\ =& \max _\pi \min _{Q\in \mathcal{P}}  \mathbb{E}\Big[  \sum_t \gamma^t  ( r(s_t,a_t) + \gamma D(P_{t;s_t,a_t}, Q_{t;s_t,a_t} )  \Big] 
\end{align*}

with $D$ a penalty function that can be chosen as KL diverengence for example and $P_{t;s_t,a_t}$ the transition kernel at time $t$ with current state action $(s_t,a_t)$.For the expectile risk measure, the corresponding $D$ is simply:

\begin{align*}
    D(P, Q)=\left\{\begin{array}{cc}0 & \text { if }  \eta \sqrt{\frac{\alpha}{1-\alpha}}\leq P(s) / Q(s) \leq  \sqrt{\frac{(1-\alpha)}{\alpha}} \eta  , \forall s \in S \\ +\infty & \text { otherwise. }\end{array}\right.
\end{align*}
where $\eta$ is defined in \ref{exextile}.
Using Theorem 2 of \cite{zhang2023regularized}, we have directly that :

 \begin{align}
         &v^*_\alpha(s)=v^\pi_{\mathcal{E}}:=\max_\pi \min _{Q \in \mathcal{E}} v^{\pi,Q}   \\
&v^\pi_\alpha(s)=v^\pi_{\mathcal{E}}:= \min _{Q \in \mathcal{E}} v^{\pi,Q}   .
    \end{align}
   
\end{proof}

\section{\texttt{AutoExpectRL} algorithm description}
\label{appendix:auto}
In the section, we gives implementation details of our algorithm \texttt{AutoExpectRL.}
First, we choose a neural network that has $4$ heads for the critic, 
 one per value of  $\alpha$, leading to $4$ estimates of the pessimist $Q$-function, $Q_{\phi_d}(s, a) ,\quad \forall d\in [1,4] $. Even if some parameters are shared in the body of the network, we denote parameters of the critic as $\phi=\{ \phi_1,\phi_2,\phi_3,\phi_4   \}$. A similar network is used for actor neural network, with four heads, one per policy $\pi_{\theta_d} , \forall d \in [1, 4].$ with $\theta=\{  
\theta_1,\theta_2,\theta_3,\theta_4 \} $.

Given $4$ target $ y_d\left(r, s^{\prime} \right)=r+\gamma Q_{{\phi_{d}}_{, \operatorname{targ}}}\left(s^{\prime}, \pi_{\theta_d}(s^{\prime})\right)$ with reward $r$, policy $\pi_{\theta_d}$, we propose to minimize the \texttt{AutoExpectRL} critic loss 

\begin{align}
    L_{auto} \left(\phi_, \mathcal{D}\right)=\underset{\left(s, a, r, s^{\prime}\right) \sim \mathcal{D}}{\mathbb{E}} \left[ \sum_{d=1}^4 L^{\alpha_d}_2\left(Q_{\phi_d}(s, a)-y_d\left(r, s^{\prime}\right)\right)\right] .
\end{align}

which as associated  \texttt{UpdateCritics}$(B,\theta,\phi$) function which is a gradient ascent using   :

\begin{equation}
\label{auto_critic}
\Delta_\phi \propto \nabla_{\phi} \frac{1}{|B|} \sum_{\left(s, a, r, s^{\prime}\right) \in B} \sum_{d=1}^4 L^{\alpha_d}_2\left(Q_{\phi_d}(s, a)-y_d\left(r, s^{\prime}\right)\right).
\end{equation}


The actor of our algorithm \texttt{AutoExpectRL} is updated according to the gradient of the sum of the actor's head losses or \texttt{UpdateActor}$( T, \theta, \phi)$:

\begin{align}
\label{auto_actor}
   \Delta \theta \propto \nabla_\theta \frac{1}{|B|} 
    \sum_{s \in B} \sum_{k=1}^4 Q_{\phi_k}\left(s, \pi_{\theta_k}(s)\right).
\end{align}

\begin{algorithm}[!t]
	\caption{\texttt{AutoExpectRL} }
  \label{autoalgo}
		\begin{algorithmic}[1]
			\STATE Initialize critic networks $Q_{\phi_d}$ and actor $\pi_\theta$ $\forall d\in [1,4]$ \\
			Initialize target networks for all networks, i.e. $\forall d \in[1,4]$ $\phi_d'\gets\phi_d$, $\theta'_d \gets\theta_d$ \\
			Initialize replay buffer and \textcolor{blue}{bandit probabilities} $\mathcal B \gets \emptyset,$ \textcolor{blue}{$\mathbf{p}_1^\alpha \gets \mathcal U([0, 1]^D)$} \\
			\FOR{episode in $m=1, 2, \dots $}
			    \STATE \textcolor{blue}{Initialize episode reward $R_m \gets 0$}
			    \STATE \textcolor{blue}{Sample expectile $\alpha_m \sim \mathbf p_m^\alpha$}
			    \FOR{time step $t=1,2,\dots,T$}
				    \STATE \textcolor{blue}{Select noisy action $a_t = \pi_{\theta_d}(s_t) + \epsilon$, $\epsilon \sim \mathcal N(0, s^2)$, obtain $r_{t+1}, s_{t+1}$ where $d$ is the index in the bandit problem of chosen expectile $\alpha_m$} 
				    \STATE \textcolor{blue}{Add to total reward $R_m \gets R_m + r_{t+1}$}
				    \STATE Store transition $\mathcal B \gets \mathcal B \cup \{(s_t, a_t, r_{t+1}, s_{t+1})\}$ 
				    \STATE Sample $N$ transitions $B = \left(s, a, r, s'\right)_{n=1}^N \sim \mathcal B$.
				    \STATE \textcolor{blue}{\texttt{Update Critics}$\left( B,\theta', \phi' \right)$} according to \eqref{auto_critic}.
				    \STATE \textbf{if} $t \mod b$ \textbf{then}
				    \STATE \qquad \textcolor{blue}{\texttt{UpdateActor}$( T, \theta, \phi)$} according to \eqref{auto_actor}.
				    \STATE \qquad Update $\phi_d'$: $\phi_d' \gets \tau \phi_d + (1 - \tau) \phi_d'$, $d\in\{1,4\}$ 
				    \STATE \qquad Update $\theta'$: $\theta'_d \gets \tau \theta_d + (1 - \tau) \theta'_d$
			    \ENDFOR
			    \STATE \textcolor{blue}{Update bandit $\mathbf p^\alpha$ weights using : $w_{m+1}(d) = w_m(d) + \eta \frac{R_m - R_{m-1}}{\mathbf{p}_m^\alpha(d)}$}
			\ENDFOR
	\end{algorithmic}
\end{algorithm}

The dimension of our neural network is related to the dimension of the classical network of TD3. First, we choose a common body of share weights for our neural network of hidden dimension $[400,300]$. Then our network is composed of $4$ heads, each with final matrix weights of dimensions $300\times1$ where $1$ represents the value of one pessimist $Q$-function $Q_k$.  The dimension of the actor-network hidden layers is similar to the critic network for share weights, but the non-shared weights between the last hidden layer and the $4$ policies have dimension $300\times \vert A\vert $. Finally, the sampling of new trajectories is done using the actor head with the chosen current  $\alpha$ proposed by the bandit algorithm using $\pi_{\theta_d}$ with $d$ the index of the chosen expectile.

The algorithm can be summarised as in Algorithm \ref{autoalgo}. The blue parts are parts that differ from the traditional TD3 algorithm, as they are related to the bandit mechanism or \texttt{ExpectRL} losses. Note that the parameter $b$, the delay between the update of the critic and the actor, is usually chosen as $2$ in TD3 algorithm. Finally, in the update of the bandit, an extra parameter, the learning rate $\eta$ of the gradient ascent must be chosen. This parameter influences how fast the bandit converges to an arm, and in our case is chosen as $0.2$ like in \cite{moskovitz2021tactical} which uses bandit to fine-tune parameters in Rl algorithm. for all environments. Finally, in the testing phase of the benchmark, the best arm is chosen to maximize the reward.

\section{Hyperparameters}
\label{appendix:hyper}
\begin{table}[ht]
    \centering
\begin{tabular}{ll}

\hline Hyperparameter & Value \\
\hline Learning rate actor & $3 \mathrm{e}-4$ \\
Learning rate critic & $3 \mathrm{e}-3$ \\
Batch size & 100 \\
Memory size & $3 \mathrm{e} 5$ \\
Gamma & 0.99 \\
Polyak update $\tau$ & 0.995 \\
Number of steps before training & $7 \mathrm{e} 4$\\
Train frequency and gradient step & 100\\
Network Hidden Layers (Critic) &[400, 300]  like original implementation of TD3\\
Network Hidden Layers (Actor) &[400, 300]  like original implementation of TD3\\
\label{hyperparam}
\end{tabular}
\caption{Hyperparameters}
\end{table}

All experiments were run on an internal cluster containing a mixture of GPU Nvidia Tesla V100 SXM2 32 Go. Each run was performed on a single GPU and lasted
between 1 and 8 hours, depending on the task and GPU model. Our baseline implementations for TD3 is \cite{stable-baselines3} where we use the same base hyperparameters across all experiments,
displayed in Table \ref{hyperparam}.

\section{\texttt{AutoExpecRL} vs  other expectiles on Robust benchmark for mean on Table~\ref{nominal} }

This section illustrates the fact that  \texttt{ExpecRL} method outperforms on robust benchmark TD3 algorithm. Without any hyperparameter tuning, \texttt{AutoExpecRL} achieves a similar performance to \texttt{ExpecRL} with the best expectile, finding the best arms in the bandit problem.
In Ant and Hopper environments, the best expectile is frequently very low, typically $\alpha=0.2 $ our $0.3$ where this is less the case for HalfCheetah and Humanoid where the best expectile is bigger. Finally, we can remark that smaller expectiles give better performance in terms of min performance while for average metric, higher expectiles are chosen, which is also verified in Table \ref{Best_expect} for DR benchmark.

\label{appendix:score}

\begin{figure}[h]
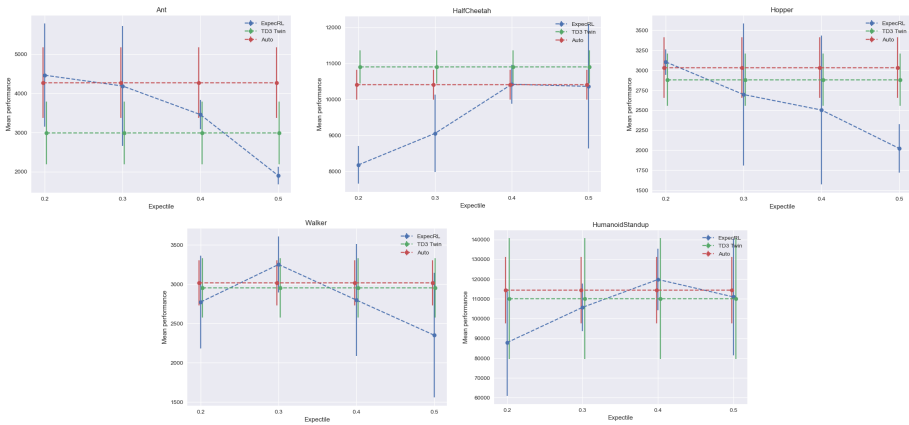

    \centering
    \begin{minipage}{.32\linewidth}
        \includegraphics[width=\linewidth]{ant_auto.pdf}
    \end{minipage}
    \hfill
    \begin{minipage}{.32\linewidth}
        \includegraphics[width=\linewidth]{half_auto.pdf}
    \end{minipage}
    \hfill
    \begin{minipage}{.32\linewidth}
        \includegraphics[width=\linewidth]{hopper_auto.pdf}
    \end{minipage}
    \hfill
    \begin{minipage}{.32\linewidth}
        \includegraphics[width=\linewidth]{walker_auto.pdf}
    \end{minipage}
    \begin{minipage}{.32\linewidth}
        \includegraphics[width=\linewidth]{hum_auto.pdf}
    \end{minipage}
    \caption{Mean performance as a function of the expectile, non-robust case (corresponding to Table~\ref{nominal}).}

\end{figure}

\begin{figure}[h]
    \centering
    \begin{minipage}{.32\linewidth}
        \includegraphics[width=\linewidth]{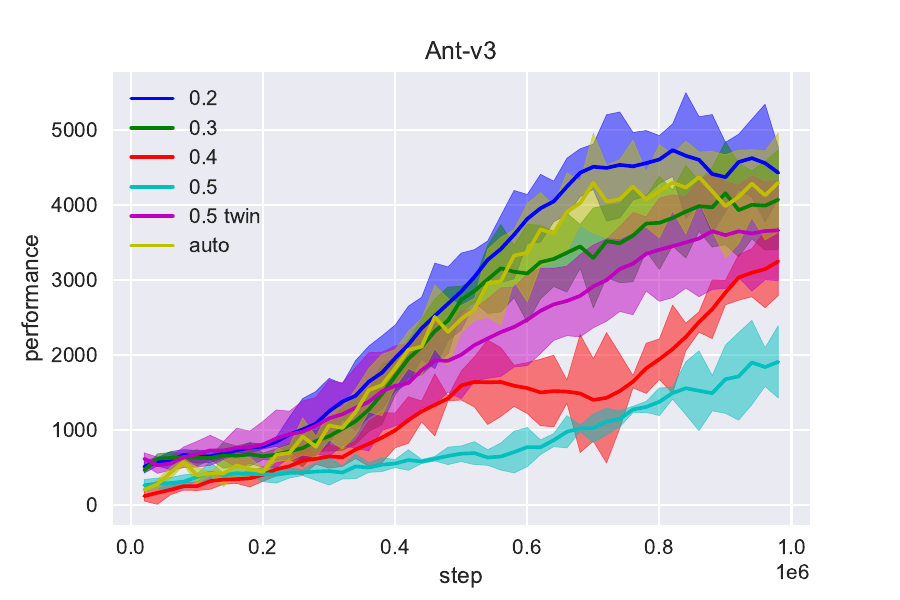}
    \end{minipage}
    \hfill
    \begin{minipage}{.32\linewidth}
        \includegraphics[width=\linewidth]{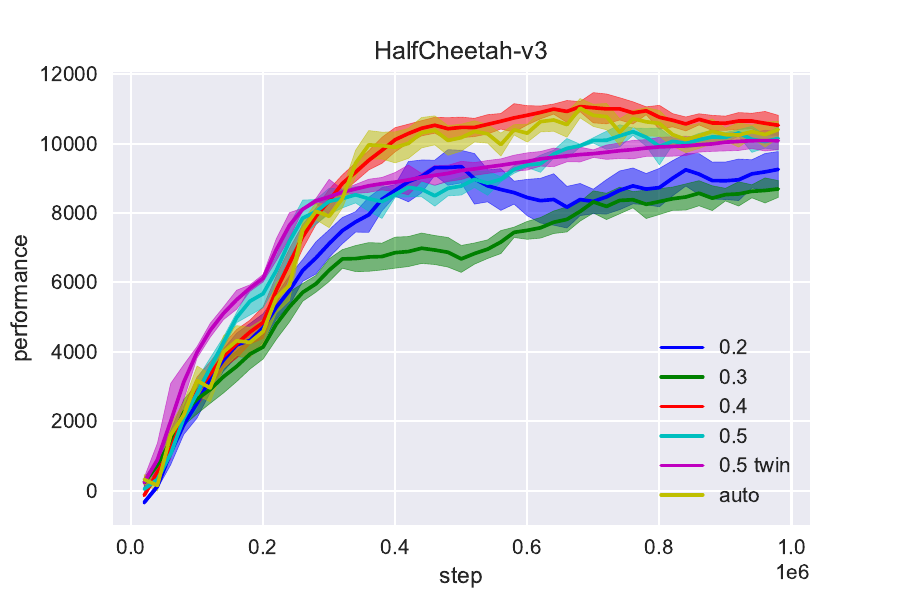}
    \end{minipage}
    \hfill
    \begin{minipage}{.32\linewidth}
        \includegraphics[width=\linewidth]{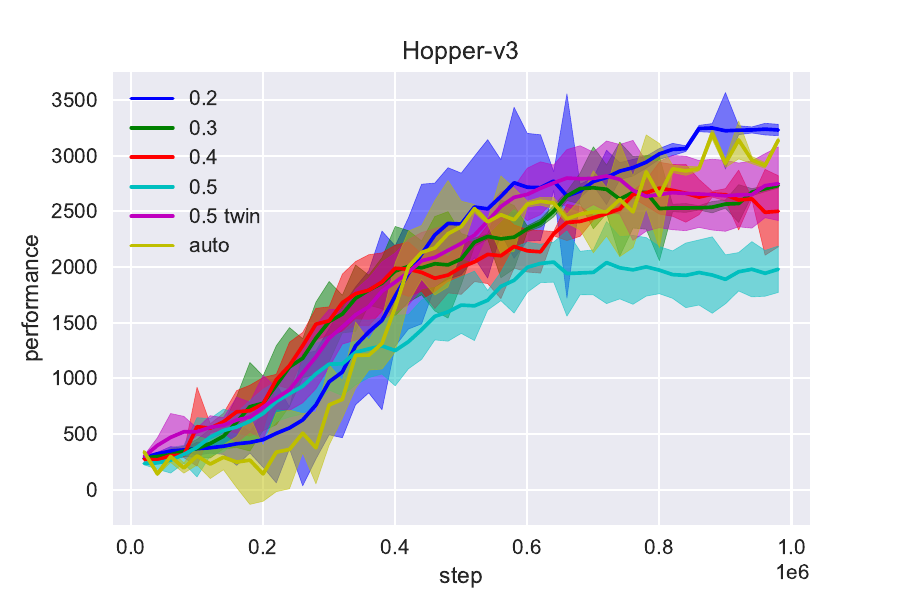}
    \end{minipage}
    \hfill
    \begin{minipage}{.32\linewidth}
        \includegraphics[width=\linewidth]{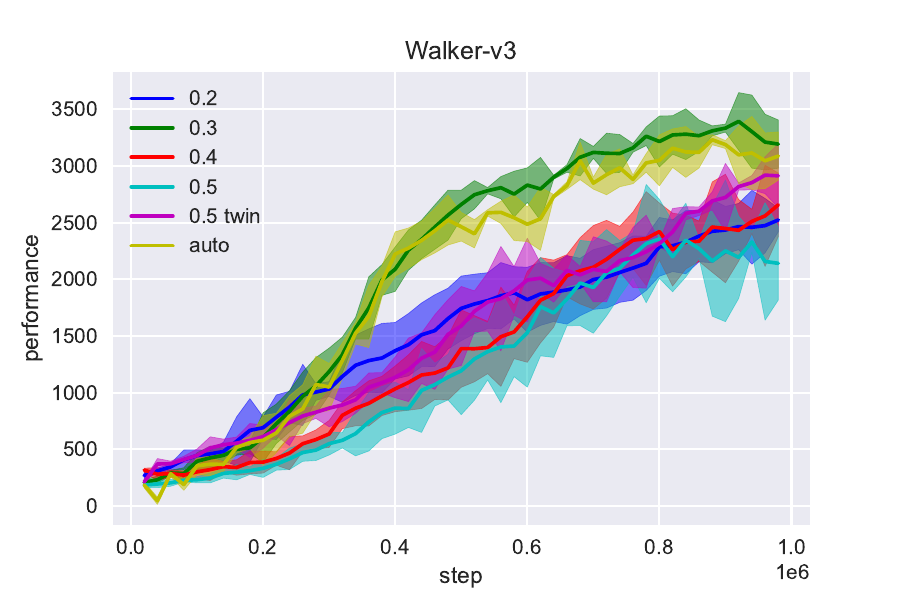}
    \end{minipage}
    \caption{Learning curves non-robust case (corresponding to Table~\ref{nominal}).}

\end{figure}

\newpage

\section{ Worst case performance for \texttt{AutoExpecRL} and \texttt{ExpecRL}  (only nominal samples) or Table \ref{classical}. }

\subsection{For 1D uncertainty greed benchmark}
\begin{figure}[h!]
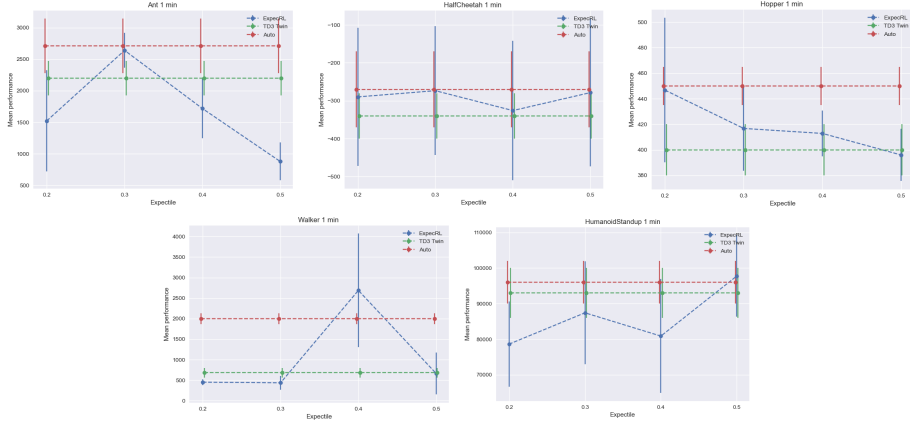

    \centering
    \begin{minipage}{.32\linewidth}
        \includegraphics[width=\linewidth]{ant1_min.pdf}
    \end{minipage}
    \hfill
    \begin{minipage}{.32\linewidth}
        \includegraphics[width=\linewidth]{half1_min.pdf}
    \end{minipage}
    \hfill
    \begin{minipage}{.32\linewidth}
        \includegraphics[width=\linewidth]{hopper1_min.pdf}
    \end{minipage}
    \hfill
    \begin{minipage}{.32\linewidth}
        \includegraphics[width=\linewidth]{walker1_min.pdf}
    \end{minipage}
    \begin{minipage}{.32\linewidth}
        \includegraphics[width=\linewidth]{hum1_min.pdf}
    \end{minipage}
    \caption{Min performance as a function of the expectile, robust case (corresponding to Table~\ref{classical}).}
    
\end{figure}

\newpage

\subsection{For 2D uncertainty greed benchmark}

\begin{figure}[h!]
    \centering
    \begin{minipage}{.32\linewidth}
        \includegraphics[width=\linewidth]{ant2_min.pdf}
    \end{minipage}
    \hfill
    \begin{minipage}{.32\linewidth}
        \includegraphics[width=\linewidth]{half2_min.pdf}
    \end{minipage}
    \hfill
    \begin{minipage}{.32\linewidth}
        \includegraphics[width=\linewidth]{hopper2_min.pdf}
    \end{minipage}
    \hfill
    \begin{minipage}{.32\linewidth}
        \includegraphics[width=\linewidth]{walker2_min.pdf}
    \end{minipage}
    \begin{minipage}{.32\linewidth}
        \includegraphics[width=\linewidth]{hum2_min.pdf}
    \end{minipage}
    \caption{Min performance as a function of the expectile, robust case (corresponding to Table~\ref{classical}).}
 
\end{figure}

\subsection{For 3D uncertainty greed benchmark}

\begin{figure}[h!]
    \centering
    \begin{minipage}{.32\linewidth}
        \includegraphics[width=\linewidth]{ant3_min.pdf}
    \end{minipage}
    \hfill
    \begin{minipage}{.32\linewidth}
        \includegraphics[width=\linewidth]{half3_min.pdf}
    \end{minipage}
    \hfill
    \begin{minipage}{.32\linewidth}
        \includegraphics[width=\linewidth]{hopper3_min.pdf}
    \end{minipage}
    \hfill
    \begin{minipage}{.32\linewidth}
        \includegraphics[width=\linewidth]{walker3_min.pdf}
    \end{minipage}
    \begin{minipage}{.32\linewidth}
        \includegraphics[width=\linewidth]{hum3_min.pdf}
    \end{minipage}
    \caption{Min performance as a function of the expectile, robust case (corresponding to Table~\ref{classical}).}
  
\end{figure}

\section{ Average performance for \texttt{AutoExpecRL} and \texttt{ExpecRL}(only nominal samples) or Table \ref{classical}. }

\newpage
\subsection{For 1D uncertainty greed benchmark}

\begin{figure}[h!]
    \centering
    \begin{minipage}{.32\linewidth}
        \includegraphics[width=\linewidth]{ant1_mean.pdf}
    \end{minipage}
    \hfill
    \begin{minipage}{.32\linewidth}
        \includegraphics[width=\linewidth]{half1_mean.pdf}
    \end{minipage}
    \hfill
    \begin{minipage}{.32\linewidth}
        \includegraphics[width=\linewidth]{hopper_1_mean.pdf}
    \end{minipage}
    \hfill
    \begin{minipage}{.32\linewidth}
        \includegraphics[width=\linewidth]{walker1_mean.pdf}
    \end{minipage}
    \begin{minipage}{.32\linewidth}
        \includegraphics[width=\linewidth]{hum1_mean.pdf}
    \end{minipage}
    \caption{Min performance as a function of the expectile, robust case (corresponding to Table~\ref{classical}).}
    
\end{figure}

\subsection{For 2D uncertainty greed benchmark}

\begin{figure}[h!]
    \centering
    \begin{minipage}{.32\linewidth}
        \includegraphics[width=\linewidth]{ant2_mean.pdf}
    \end{minipage}
    \hfill
    \begin{minipage}{.32\linewidth}
        \includegraphics[width=\linewidth]{half2_mean.pdf}
    \end{minipage}
    \hfill
    \begin{minipage}{.32\linewidth}
        \includegraphics[width=\linewidth]{hopper2_mean.pdf}
    \end{minipage}
    \hfill
    \begin{minipage}{.32\linewidth}
        \includegraphics[width=\linewidth]{walker2_mean.pdf}
    \end{minipage}
    \begin{minipage}{.32\linewidth}
        \includegraphics[width=\linewidth]{hum2_mean.pdf}
    \end{minipage}
    \caption{Min performance as a function of the expectile, robust case (corresponding to Table~\ref{classical}).}
   
\end{figure}

\newpage
\subsection{For 3D uncertainty greed benchmark}

\begin{figure}[h!]
    \centering
    \begin{minipage}{.32\linewidth}
        \includegraphics[width=\linewidth]{ant3_mean.pdf}
    \end{minipage}
    \hfill
    \begin{minipage}{.32\linewidth}
        \includegraphics[width=\linewidth]{half3_mean.pdf}
    \end{minipage}
    \hfill
    \begin{minipage}{.32\linewidth}
        \includegraphics[width=\linewidth]{hopper3_mean.pdf}
    \end{minipage}
    \hfill
    \begin{minipage}{.32\linewidth}
        \includegraphics[width=\linewidth]{walker3_mean.pdf}
    \end{minipage}
    \begin{minipage}{.32\linewidth}
        \includegraphics[width=\linewidth]{hum3_mean.pdf}
    \end{minipage}
    \caption{Min performance as a function of the expectile, robust case (corresponding to Table~\ref{classical}).}
   
\end{figure}

    

  

    

    

    

\section{ Additional details for expectiles on Robust benchmark for worst-case and mean on Table \ref{benchmark:robust}  }

\begin{table*}[h!]
\begin{center}
\scriptsize
\begin{tabular}{|l|l|l|} \hline   Env & Min & Mean 
\\  \hline \hline $Ant1 $  & 3  & 3
\\ \hline $Ant2 $ & 2 & 3
\\ \hline $Ant3  $ &  2 & 3 
\\ \hline $HalfCheetah 1 $ & 3 &3
\\ \hline $HalfCheetah 2 $ &  3 &3
\\ \hline $HalfCheetah 3  $ & 3 &3 
\\ \hline $Hopper 1 $ & 3  & 4
\\ \hline $Hopper 2 $ &  3 & 4
\\ \hline $Hopper 3  $ & 3  & 3
\\ \hline $Walker 1 $ &  3  & 4
\\ \hline $Walker 2 $ & 4  & 4 
\\ \hline $Walker 3 $ & 3  &3 
\\  \hline $HumanoidStandup 1 $ &  3 &3
\\  \hline $HumanoidStandup 2 $ &  2 &3
\\ \hline $HumanoidStandup 3 $ &  2  &3
\\ \hline
\end{tabular}
\caption{Best Expectile in  DR for \texttt{ExpectRL}  } 
\label{Best_expect}
\end{center}

\end{table*}

\newpage

\section{Uncertainty sets used for Robust benchmark}

\label{appendix:uncertainty}

\begin{table*}[h!]
\centering
\caption{Uncertainty sets used for Robust benchmark}
\label{table:environment_settings}
\fontsize{6}{10}\selectfont
\begin{tabular}{|l|l|l|l|}
\bottomrule
Environment          & Uncertainty Set $\Omega$                         & Reference Parameter & Uncertainty Parameter Name                                 \\ \toprule\bottomrule
\multicolumn{4}{|l|}{Baseline MuJoCo Environment: Ant}\\ \hline
Ant 1                 & {[}0.1, 3.0{]}                                    & 0.33                & torso mass
 \\ \hline
Ant 2                 & {[}0.1, 3.0{]} $\times$ {[}0.01, 3.0{]}                 & (0.33, 0.04)         & torso mass $\times$ front left leg mass
 \\ \hline
Ant 3                 & {[}0.1, 3.0{]} $\times$ {[}0.01, 3.0{]} $\times$ {[}0.01, 3.0{]} & (0.33, 0.04, 0.06) & torso mass $\times$ front left leg mass $\times$ front right leg mass
 \\ \toprule \bottomrule
\multicolumn{4}{|l|}{Baseline MuJoCo Environment: HalfCheetah}\\ \hline
HalfCheetah 1        & {[}0.1, 4.0{]}                                    & 0.4                 & world friction                                 \\ \hline
HalfCheetah 2        & {[}0.1, 4.0{]} $\times$ {[}0.1, 7.0{]}                  & (0.4, 6.36)          & world friction $\times$ torso mass                    \\ \hline
HalfCheetah 3        & {[}0.1, 4.0{]} $\times$ {[}0.1, 7.0{]} $\times$ {[}0.1, 3.0{]}  & (0.4, 6.36, 1.53)   & world friction $\times$ torso mass $\times$ back thigh mass  \\ \toprule\bottomrule
\multicolumn{4}{|l|}{Baseline MuJoCo Environment: Hopper}\\ \hline
Hopper 1              & {[}0.1, 3.0{]}                                    & 1.00                & world friction                                 \\ \hline
Hopper 2              & {[}0.1, 3.0{]} $\times$ {[}0.1, 3.0{]}                   & (1.00, 3.53)         & world friction $\times$ torso mass                    \\ \hline
Hopper 3              & {[}0.1, 3.0{]} $\times$ {[}0.1, 3.0{]} $\times$ {[}0.1, 4.0{]}  & (1.00, 3.53, 3.93)  & world friction $\times$ torso mass $\times$ thigh mass       \\ \toprule\bottomrule
\multicolumn{4}{|l|}{Baseline MuJoCo Environment: HumanoidStandup}\\ \hline
HumanoidStandup 1    & {[}0.1, 16.0{]}                                   & 8.32                & torso mass                                     \\ \hline
HumanoidStandup 2    & {[}0.1, 16.0{]} $\times$ {[}0.1, 8.0{]}                  & (8.32, 1.77)         & torso mass $\times$ right foot mass                   \\ \hline
HumanoidStandup 3    & {[}0.1, 16.0{]} $\times$ {[}0.1, 5.0{]} $\times$ {[}0.1, 8.0{]} & (8.32, 1.77, 4.53)  & torso mass $\times$ right foot mass $\times$ left thigh mass \\ \toprule\bottomrule
\multicolumn{4}{|l|}{Baseline MuJoCo Environment: Walker}\\ \hline
Walker 1             & {[}0.1, 4.0{]}                                    & 0.7                & world friction                                 \\ \hline
Walker 2             & {[}0.1, 4.0{]} $\times$ {[}0.1, 5.0{]}                   & (0.7, 3.53)          & world friction $\times$ torso mass                    \\ \hline
Walker 3             & {[}0.1, 4.0{]} $\times$ {[}0.1, 5.0{]} $\times$ {[}0.1, 6.0{]}  & (0.7, 3.53, 3.93)& world friction $\times$ torso mass $\times$ thigh mass       \\ \toprule \bottomrule

\end{tabular}

\end{table*}

\newpage

\end{document}